\documentclass[journal]{IEEEtran}
\usepackage{diagbox}

\usepackage[ruled,linesnumbered]{algorithm2e}
\usepackage{amsmath,amssymb,amsfonts}
\usepackage{graphicx}
\usepackage{epstopdf}
\usepackage{subfig}
\usepackage{cite}
\usepackage{gensymb}
\usepackage{multirow,booktabs,color,soul,threeparttable,graphicx}
\newcommand{\RNum}[1]{\uppercase\expandafter{\romannumeral #1\relax}}

\newcommand{\ieeepreprintnotice}{%
This work has been submitted to the IEEE for possible publication. Copyright may be transferred without notice, after which this version may no longer be accessible.%
}

\definecolor{hl}{rgb}{0.75,0.75,0.75}
\sethlcolor{hl}
\definecolor{emph}{rgb}{0,0,1}
\ifCLASSINFOpdf

\else

\fi

\begin{document}

\title{Relation Reasoning with LLMs in Expensive Optimization\thanks{\ieeepreprintnotice}}
 
\author{Ye~Lu, Bingdong~Li, Aimin~Zhou, Hao~Hao
\thanks{Y. Lu, b. Li, A. Zhou and H. Hao are with the Shanghai Institute of Al for Education, East China Normal University, Shanghai 200062, China (Corresponding author: Hao Hao; e-mail: hhao@mail.ecnu.edu.cn).}
}

\maketitle

\begin{abstract}
Expensive optimization problems (EOPs) are black-box tasks with costly objective evaluations and no gradient access, making the evaluation budget the key bottleneck. Surrogate-assisted evolutionary algorithms (SAEAs) reduce evaluations via surrogate predictions, but conventional surrogates often require frequent retraining as populations evolve, incurring overhead. This paper proposes R2SAEA, a reinforcement-trained relation-based large language model (LLM) surrogate assisted evolutionary algorithm. We cast relation-based surrogate modeling as an in-context pairwise reasoning task. To enable efficient inference in evolutionary loops, we develop an anchor-based iterative context construction strategy that reduces prompt complexity from quadratic to linear in population size, and a voting-based aggregation scheme that converts predicted relations into scores for offspring selection. We further build an RL pipeline from evolutionary trajectories and fine-tune Qwen2.5 with GRPO. Experiments on single- and multi-objective benchmarks show improved relation prediction and state-of-the-art optimization performance over strong SAEA baselines and general LLMs. Quantization also enables efficient edge deployment, supporting a zero-shot surrogate paradigm without per-generation retraining. Code and models are available at \emph{https://github.com/Septend9/R2SAEA}.
\end{abstract}


\IEEEpeerreviewmaketitle

\section{Introduction}

Expensive optimization problems (EOPs)~\cite{li2022evolutionary} are ubiquitous in science and engineering, where each objective evaluation may require physical experiments, high-fidelity simulations, or time-consuming computations~\cite{9508774}. These problems are typically black-box: the evaluator is opaque, gradients are unavailable, and optimization must rely solely on input--output observations. As a consequence, the evaluation budget becomes the dominant bottleneck, and the central goal is to obtain high-quality solutions under a severely limited number of expensive evaluations.

Evolutionary algorithms (EAs) are a major family of derivative-free optimizers for black-box settings because of their robustness and minimal assumptions~\cite{yu2016derivative}. However, vanilla EAs usually require a large number of evaluations, which is often infeasible for EOPs. SAEAs~\cite{jin2011surrogateassisted} alleviate this difficulty by learning inexpensive models from previously evaluated solutions and using them to screen or rank newly generated candidates. Over the past decades, most surrogate models in SAEAs have been designed following three paradigms~\cite{liu2025dual}: regression (predicting objective values), classification (predicting quality labels), and relation modeling (predicting pairwise superiority). Among them, relation models are particularly appealing because EAs are inherently comparison-driven: selection operators depend primarily on relative superiority rather than absolute objective values~\cite{loshchilov2010comparison}. Despite their success, conventional surrogates in SAEAs commonly follow an online ``train--predict'' loop, where the surrogate is retrained frequently to track the continuously shifting population distribution. Such repeated training can impose nontrivial overhead, and in some scenarios it undermines the efficiency gains brought by surrogate assistance.

The rapid progress of large language models (LLMs) has created new opportunities for surrogate modeling. Recent studies have explored using general-purpose LLMs as regression or classification surrogates via prompting~\cite{hao2024largelanguagemodelssurrogate}. However, directly deploying frontier LLMs in evolutionary loops is often impractical due to high inference cost and latency, while smaller models may lack the task-specific reasoning capability to reliably infer subtle superiority relations among solutions. More importantly, although relation modeling is a natural fit to EAs, LLMs have not yet been systematically studied as relation surrogates in SAEAs, especially under strict token budgets and repeated inference calls.

This paper aims to bridge these gaps by developing an LLM surrogate that is (1) specialized for relation prediction, (2) efficient enough for repeated use in evolutionary optimization, and (3) deployable without per-generation retraining. We propose the Reinforcement-trained relation-based LLM surrogate assisted evolutionary algorithm (R2SAEA). Our key idea is to cast relation surrogate modeling as an in-context reasoning task and to endow a compact LLM with robust relation inference ability via reinforcement learning fine-tuning. Concretely, We adopt two mainstream relation-surrogate criteria, the function criterion (C1) and the category criterion (C2)~\cite{hao2025expensive}, and design a prompt template that supports both scenarios. To make in-context inference scalable, we further introduce an anchor-based iterative context construction strategy that decomposes quadratic pairwise comparisons into a sequence of linear-sized prompts. The predicted pairwise relations are then aggregated into absolute quality indicators via a voting-based scoring scheme, enabling efficient offspring selection.

To strengthen relation reasoning while avoiding the cost of frontier models, we build an RL fine-tuning pipeline from evolutionary trajectories and train Qwen2.5 using Group Relative Policy Optimization (GRPO)~\cite{shao2024deepseekmath}. The resulting relation LLM surrogates exhibit strong zero-shot generalization across benchmark functions and optimization stages, and they can be embedded into standard SAEA frameworks without any online retraining. In addition, by quantizing compact variants, we demonstrate the feasibility of deploying relation LLM surrogates on edge devices, pointing to a practical hardware-level surrogate paradigm for expensive optimization.

The main contributions of this work are summarized as follows:
\begin{enumerate}
    \item We formulate relation-based surrogate modeling as an in-context LLM reasoning problem under two mainstream criteria (C1/C2), and propose an anchor-based iterative prompt construction mechanism to reduce context complexity.
    \item We design a voting-based aggregation strategy to convert predicted pairwise relations into absolute scores for candidate ranking and selection in SAEAs.
    \item We build an end-to-end RL fine-tuning pipeline from evolutionary trajectories and train Qwen2.5 with GRPO, obtaining relation LLM surrogates that outperform strong general-purpose LLMs on relation prediction benchmarks; quantized models further enable efficient edge deployment without per-generation retraining.
    \item By embedding the trained surrogates into EAs, we achieve state-of-the-art performance on both single- and multi-objective expensive optimization benchmarks.
\end{enumerate}

The remainder of this paper is organized as follows. Section~\ref{sec:preliminaries} reviews SAEAs, relation modeling, LLM-based optimization, and RL fine-tuning. Section~\ref{sec:proposed_method} presents the proposed R2SAEA framework, including context engineering, voting-based scoring, and GRPO training. Section~\ref{sec:experiments} reports experimental evaluations and ablation studies. Section~\ref{sec:conclusion} concludes the paper and outlines future directions.

\section{Preliminaries}
\label{sec:preliminaries}

This section reviews four related directions: surrogate-assisted evolutionary optimization, relation models, LLM--evolutionary algorithm integration, and reinforcement-learning-based fine-tuning for task adaptation.

\subsection{Surrogate assisted Evolutionary Algorithms}
\label{subsec:preliminaries-SAEA}

Surrogate-assisted evolutionary algorithms (SAEAs) are typically studied for expensive black-box optimization, where the objective function(s) are unknown, non-differentiable, or only accessible through costly evaluations~\cite{wang2026review}. This setting can be expressed as
\begin{equation}
\min_{\mathbf{x}\in \mathcal{X}} \ \mathbf{f}(\mathbf{x}),
\end{equation}
where $\mathbf{x}\in\mathbb{R}^{D}$ is the decision vector, $\mathcal{X}\subseteq \mathbb{R}^{D}$ denotes the feasible region, and $\mathbf{f}(\mathbf{x})$ is expensive to evaluate. For single-objective optimization, $\mathbf{f}(\mathbf{x})=f(\mathbf{x})\in\mathbb{R}$ and the goal is to find $\mathbf{x}^{*}=\arg\min_{\mathbf{x}\in\mathcal{X}} f(\mathbf{x})$. For multi-objective optimization, $\mathbf{f}(\mathbf{x})=(f_{1}(\mathbf{x}),\ldots,f_{M}(\mathbf{x}))$ and solutions are compared by Pareto dominance, with the aim of approximating the Pareto-optimal set (and its corresponding Pareto front)~\cite{zhou2011multiobjective}.

Evolutionary algorithms are gradient-free and thus suitable for black-box problems, but they often require many expensive evaluations. SAEAs mitigate this by training surrogates on evaluated solutions to approximate the objective(s), guiding search with fewer real evaluations. A central challenge in designing SAEAs lies in constructing effective surrogate models and, crucially, in developing corresponding model-management strategies. Jin et al.~\cite{jin2011surrogateassisted} provide a comprehensive review of model management, which enables surrogates to be more seamlessly and reliably integrated into evolutionary search. In parallel, surrogate modeling in SAEAs has gradually converged into three major paradigms~\cite{liu2025dual}: regression-, classification-, and relation-based approaches.

In regression-based SAEAs, the expensive objective function is approximated by a curve fitted to sampled data points; representative models include polynomial regression~\cite{lian2005multiobjective}, radial basis function (RBF) networks~\cite{sun2017surrogate}, and Gaussian processes (GPs)~\cite{liu2013gaussian}. Classification-based SAEAs, by contrast, assign quality labels to candidate solutions and learn decision boundaries using models such as support vector machines (SVM)~\cite{hao2021approximated}, artificial neural networks (ANN)~\cite{pan2018classification}, and fuzzy $k$-nearest neighbor (KNN)~\cite{zhou2019fuzzy}. Relation-based surrogates model the relative relationships among solutions rather than their absolute objective values~\cite{wang2026review}. They have attracted growing attention in recent years because this perspective aligns more naturally with the selection-driven essence of evolutionary algorithms. Further details are provided in the following subsections.

\subsection{Relation Learning and Prediction}
\label{subsec:preliminaries-Rela}

Relation models~\cite{hao2025expensive} are a class of surrogate model construction strategies that directly learn the comparative advantages between solutions. The fundamental concept is as follows: the model \( \mathcal{R}(\cdot)\) is designed to learn the relative advantages among solutions by direct assessment. The relation model is formally defined as:
\begin{equation}
l = \mathcal{R}(\mathbf{x}_1, \mathbf{x}_2)
\label{eq:relation}
\end{equation}
In this equation, \( \mathcal{R} \) represents the relation model, while \( \mathbf{x}_1 \) and \( \mathbf{x}_2 \) denote two solution vectors, and \( r \) signifies the predicted relationship between these solutions. As an emerging category of surrogate models, there has been rapid development in this area recently. For instance, Hao et al. ~\cite{hao2020binary,hao2024enhancing} have leveraged direct relation learning and prediction techniques to enhance the performance of differential evolution and estimation of distribution algorithms in single-objective optimization problems. In the realm of multi-objective optimization, dominance relationships between solutions are directly integrated into the construction of surrogate models, thereby boosting algorithm performance. This approach is supported by the works of Yuan et al.~\cite{yuan2021expensive}, Hao et al.~\cite{hao2022expensive}, and Tian et al.~\cite{tian2017platemo}. Recently, Hao et al.~\cite{hao2025expensive} proposed a unified framework for relation modeling. It seeks to standardize how relationships among solutions are represented, thereby offering a more coherent foundation for constructing relation-based surrogate models. Building on this line of research, Liu et al.~\cite{liu2025dual} further applied relation models to multi-objective constrained optimization. The present work likewise proceeds within the relation-model framework.

\subsection{Large Language Models for Evolutionary Optimization}
\label{subsec:preliminaries-LLM4EA}

In recent years, the fusion of LLMs and evolutionary computation has emerged as a vibrant research frontier. Liu et al.~\cite{wu2024evolutionary} investigate the collaborative potential of integrating LLMs with EAs, showing that LLMs can enhance optimization via their predictive capabilities, while EAs can, in turn, improve LLM performance through powerful optimization mechanisms~\cite{yang2024large,liu2024evolution}. Liu et al.~\cite{wu2024evolutionary} highlight the complementary strengths of integrating LLMs with evolutionary algorithms (EAs): LLMs can support optimization through prediction and generation, while EAs can refine LLMs and their outputs via iterative search~\cite{liu2023algorithm}. This synergy has been demonstrated in applications such as neural architecture search~\cite{wu2024evolutionary} and Bayesian optimization~\cite{liu2024large}. Existing studies mainly fall into two threads. For solution generation, LLMs are used to propose candidates for evolutionary optimization~\cite{meyerson2023language}, and have even been framed as evolution strategies~\cite{lange2024large}. For algorithm generation, LLMs assist in designing or improving optimization algorithms for tasks such as the traveling salesman problem and automated adversarial attacks~\cite{wu2023llm,guo2024autoda}, with systems such as OptiMUS~\cite{ahmaditeshnizi2023optimus} and OpenELM~\cite{bradley2024openelm} further illustrating this automation trend.

Despite this progress, using LLMs as surrogate models remains a nascent direction. Hao et al. leverage the general reasoning capabilities of foundation models to build surrogates for classification and regression tasks, yet their approach is still constrained by inference cost and model capacity~\cite{hao2024large}. This work seeks to mitigate these limitations through model fine-tuning.

\subsection{Reinforcement Learning for LLMs}
\label{subsec:preliminaries-RL}

With the rapid development of LLMs, reinforcement learning has become central to aligning model behavior with human intent and improving complex reasoning~\cite{christiano2017deep}. In RLHF~\cite{stiennon2020learning}, PPO~\cite{schulman2017proximal} remains a widely used policy-optimization method due to its stability, but it typically requires a critic for value estimation, increasing memory and computation and potentially inducing instability from value bias. GRPO~\cite{shao2024deepseekmath} mitigates these issues by estimating advantages via within-group normalization, eliminating the need to train an explicit critic.

Reinforcement learning is increasingly being applied beyond general-purpose settings to specialized domains. In finance, Fin-R1 adapts GRPO to improve reasoning for financial statement analysis and risk assessment~\cite{liu2025finr1largelanguagemodel}. In law, RL-based methods have been used to enhance performance in legal decision-making~\cite{dai2025legaldeltaenhancinglegalreasoning}, statutory reasoning~\cite{Hu_2025}, and legal calculation~\cite{zhang2025legalmathematicalreasoningllms}. In education, RL supports automatic optimization and personalization of teaching strategies~\cite{chu2025llmagentseducationadvances}; EduAlign further combines GRPO to align tutoring behavior with educational goals, emphasizing value alignment, individualized feedback, and creative guidance~\cite{song2025cultivatinghelpfulpersonalizedcreative}. Building on these advances, we apply RL to expensive optimization by integrating prompt engineering with GRPO to strengthen LLM relation reasoning for black-box optimization tasks.

\section{Proposed method}
\label{sec:proposed_method}
\begin{figure*}[ht!]
    \centering
    \includegraphics[width=1\linewidth]{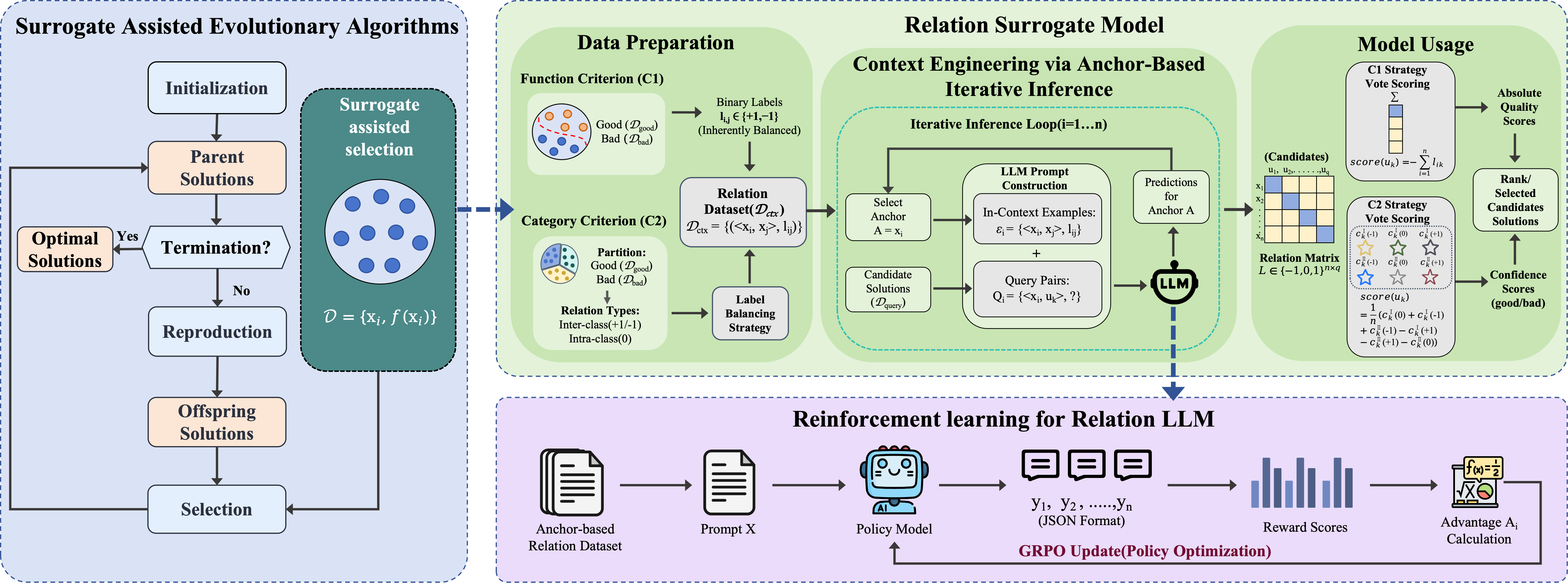}
    \caption{Overview of the R2SAEA framework. It comprises three components: an SAEA module, an LLM-driven relation model, and an end-to-end reinforcement learning module. The left panel shows a generic SAEA loop of reproduction and selection, where surrogate assistance reduces per-iteration cost. The relation model includes data preparation, context engineering, and model usage for learning and predicting relation pairs. The lower panel shows RL fine-tuning of the LLM for relational learning.}
    \label{fig:framework}
\end{figure*}

This section presents a general framework in which LLM driven relation surrogate assisted EAs in solving expensive single- and multi-objective optimization problems, as illustrated in Fig.~\ref{fig:framework}. The left panel shows an SAEA framework, comprising population initialization, the generation of new solutions, model-assisted selection, and the stopping criterion. The relation model framework consists of three stages~\cite{hao2025expensive}: data preparation, model training (implemented in this work via context engineering), and model usage.  It further supports a function criterion (C1) and a category criterion (C2), which constitute two general paradigms for relational surrogate modeling. The reinforcement learning module includes data generation, reward design, and model training. The following sections detail the specific implementation of each component.

\subsection{Relation Reasoning Task Description}
\label{subsec:proposed_method_TD}

Following the relation-learning framework of Hao et al.~\cite{hao2025expensive}, we adopt a three-stage pipeline: data preparation, in-context inference, and relation aggregation. Specifically, the original evaluated set $\mathcal{D}=\{(\mathbf{x}_i,f(\mathbf{x}_i))\}$ is transformed into a pairwise dataset $\mathcal{D}_r=\{(\langle \mathbf{x}_i,\mathbf{x}_j\rangle,l_{i,j})\}$, where $l_{i,j}$ denotes the relation between $\mathbf{x}_i$ and $\mathbf{x}_j$. The LLM then infers and predicts $l_{i,j}$ from the provided context, and the predicted relations are aggregated into scalar quality scores using established rules.

\subsubsection{Data preparation}
To train a relation model, we transform the original data samples into a set of relation pairs. Two primary strategies are employed for this data preparation: the function criterion~(C1) and the category criterion~(C2).

The function criterion~(C1) directly compares pairs of solutions $(\mathbf{x}_i, \mathbf{x}_j)$ and assigns a binary label $l_{i,j} \in \{+1, -1\}$ based on the superiority of their objective function values. For instance, $l_{i,j} = +1$ if solution $\mathbf{x}_i$ is better than $\mathbf{x}_j$. For multi-objective problems (MOPs), this is extended either by using multi-label outputs (one for each objective)~\cite{hao2021approximated} or by first scalarizing the objective vectors into a single fitness value. 

The category criterion~(C2) first partitions the dataset $\mathcal{D}$ into distinct subsets, such as good ($\mathcal{D}_{\text{good}}$) and bad ($\mathcal{D}_{\text{bad}}$) solutions, based on a specific rule (e.g., superiority or inferiority in  SOPs and Pareto dominance in MOPs). Relation pairs are then labeled to distinguish inter-class relationships (e.g., ``+1'' for a pair from $\mathcal{D}_{\text{good}}$ and $\mathcal{D}_{\text{bad}}$) from intra-class relationships (``0'' for a pair from the same subset). As this typically results in an imbalanced number of relation types, a label balancing strategy, such as random under-sampling of the majority classes, is applied to ensure equal representation of inter-class and intra-class samples in the final training set $\mathcal{D}_r$.

These two strategies enable the transformation of the original dataset into either a binary or ternary classification dataset. The assigned labels reflect the superiority or inferiority relationship between pairs of samples.

\subsubsection{Context Engineering via Anchor-Based Iterative Inference}
\label{subsec:AnchorBasedEngineering}

Our approach differs from conventional surrogates in that both learning and prediction are performed in-context by an LLM, rather than through an explicit train--predict loop. Specifically, we formulate relation prediction as an in-context pairwise reasoning task. Let $\mathcal{D}_{\mathrm{ctx}}$ denote the set of truly evaluated solutions used as context, and let $\mathcal{D}_{\mathrm{query}}$ denote the set of candidate solutions whose quality are to be predicted.

\begin{figure}[ht!]
    \centering
    \includegraphics[width=0.8\linewidth]{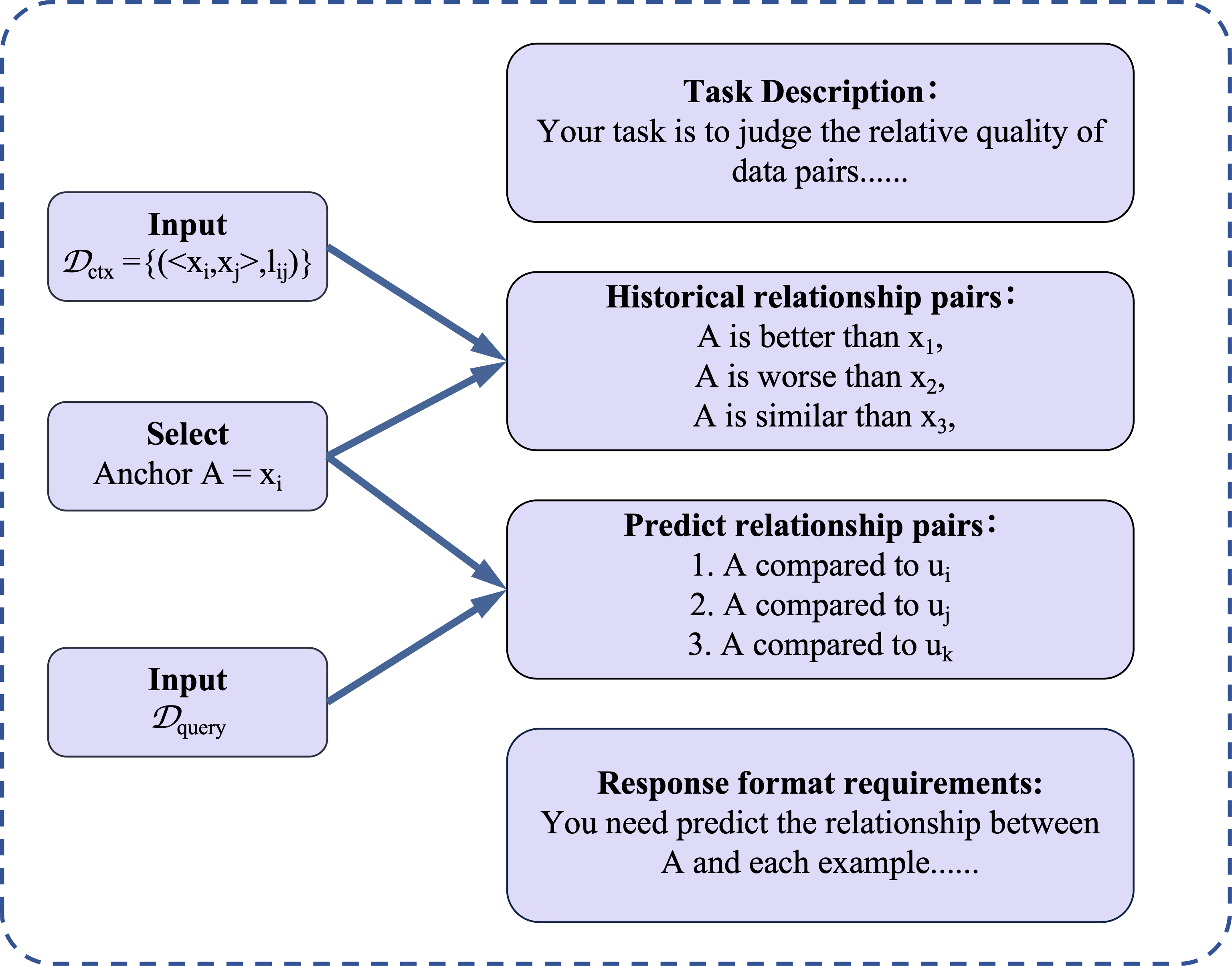}
    \caption{Prompt template for relation reasoning.}
    \label{fig:prompt_template}
\end{figure}

Each prompt consists of: (1) a brief task description, (2) a set of labeled in-context examples sampled from $\mathcal{D}_{\mathrm{ctx}}$, (3) a set of unlabeled query pairs constructed from $\mathcal{D}_{\mathrm{query}}$, and (4) a strict output format specification. Figure~\ref{fig:prompt_template} illustrates the prompt template.

A naive construction would enumerate all pairwise relations within $\mathcal{D}_{\mathrm{ctx}}\cup\mathcal{D}_{\mathrm{query}}$. Let $|\mathcal{D}_{\mathrm{ctx}}|=n$ and $|\mathcal{D}_{\mathrm{query}}|=q$. This yields $O\!\left((n+q)^2\right)$ pairs, which raises two practical issues: (1) the prompt can easily exceed the context-length limit; and (2) combining a large number of labeled relations and unlabeled queries in a single call increases the reasoning burden and may degrade prediction accuracy.

To address these issues, we propose an anchor-based iterative context construction strategy. Instead of issuing one large prompt, we decompose the task into $n$ smaller inference rounds, each centered on one context solution as a fixed reference (anchor). This reduces the per-prompt pair complexity from quadratic to linear and makes the comparison criterion more stable. The procedure is as follows:
\begin{enumerate}
    \item \textbf{Anchors.} For each context solution $\mathbf{x}_i \in \mathcal{D}_{\mathrm{ctx}}$ ($i=1,\dots,n$), we set the anchor $\mathbf{x}_i$ and perform one inference round.
  
    \item \textbf{In-context examples (labeled).} For anchor $\mathbf{x}_i$, we form labeled example pairs by pairing it with all other context solutions in $\mathcal{D}_{\mathrm{ctx}}$:
    \begin{equation}
        \mathcal{E}_i=\left\{\left(\left\langle \mathbf{x}_i, \mathbf{x}_j \right\rangle,\, l_{ij}\right)\right\}_{j=1,\,j\neq i}^{n},
    \end{equation}
    where $l_{ij}$ is the ground-truth relation label between $\mathbf{x}_i$ and $\mathbf{x}_j$.
  
    \item \textbf{Query pairs (unlabeled).} We form query pairs by pairing the same anchor with each candidate solution $\mathbf{u}_k\in\mathcal{D}_{\mathrm{query}}$:
    \begin{equation}
        \mathcal{Q}_i=\left\{\left\langle \mathbf{x}_i, \mathbf{u}_k \right\rangle\right\}_{k=1}^{q}.
    \end{equation}
    \item \textbf{Iterative inference.} We issue one prompt for each anchor $\mathbf{x}_i$ ($i=1,\ldots,n$). The prompt contains $(n\!-\!1)$ labeled context pairs from $\mathcal{E}_i$ and $q$ unlabeled query pairs from $\mathcal{Q}_i$. The LLM returns the $q$ relation labels for $\mathcal{Q}_i$ in a single completion. Therefore, each prompt contains $(n-1)+q$ pairs, reducing the per-prompt pair complexity from $O((n+q)^2)$ to $O(n+q)$. Across all $n$ anchors, we obtain $n$ prompts and a total of $nq$ predicted relations between $\mathcal{D}_{\mathrm{ctx}}$ and $\mathcal{D}_{\mathrm{query}}$.
\end{enumerate}

This anchor-based iterative framework alleviates context-length overflow and improves inference stability.

Before constructing the relational prompts, each decision vector is normalized dimension-wise to $[0,1]$ and rounded to $\beta$ decimal places (default $\beta=5$) to improve numerical stability and reduce token length.

The general prompt template used to construct our relational data is shown in Fig.~\ref{fig:prompt_template}. For the C1 criterion, we adopt a binary relation labeling scheme to indicate relative superiority: ``+1'' means that the first solution is superior to the second (better), whereas ``-1'' indicates the opposite (worse). For the C2 criterion, we use a ternary relation labeling scheme to capture both inter-class and intra-class relationships. In particular, ``+1'' and ``-1'' represent inter-class superiority and inferiority, respectively, while ``0'' indicates that the two solutions fall into the same category (similar). The full prompt specifications for C1 and C2 are provided in the supplementary material.

\subsubsection{Model Usage}
\label{subsec:model_usage}

The anchor-based iterative inference produces $n$ rounds of predictions, one round for each anchor $\mathbf{x}_i\in\mathcal{D}_{\mathrm{ctx}}$ ($i=1,\ldots,n$), over the same set of unevaluated solutions $\{\mathbf{u}_1,\ldots,\mathbf{u}_q\}\subset\mathcal{D}_{\mathrm{query}}$. Each round returns the relations between the anchor and all candidates, i.e., $\{l_{ik}\}_{k=1}^{q}$. Aggregating all rounds yields $nq$ pairwise relation predictions between $\mathcal{D}_{\mathrm{ctx}}$ and $\mathcal{D}_{\mathrm{query}}$. We then convert these pairwise relations into an absolute quality score for each candidate $\mathbf{u}_k$, which can be directly used for ranking and offspring selection in the SAEA.

We interpret the $n$ anchors as $n$ ``experts'' that collectively judge each candidate. Let
\begin{equation}
\mathbf{L}=\big[l_{ik}\big]\in\mathbb{Z}^{n\times q}
\end{equation}
denote the relation matrix predicted by the LLM, where $l_{ik}$ represents the predicted relation between anchor $\mathbf{x}_i$ and candidate $\mathbf{u}_k$.

\paragraph{C1 voting}
For the C1 criterion, $\mathbf{L}\in\{-1,+1\}^{n\times q}$, where $l_{ik}=+1$ indicates $\mathbf{x}_i$ is better than $\mathbf{u}_k$, and $l_{ik}=-1$ indicates $\mathbf{u}_k$ is better than $\mathbf{x}_i$. We define the vote score of $\mathbf{u}_k$ as the negative column sum:
\begin{equation}
\mathrm{score}_{C1}(\mathbf{u}_k)=-\sum_{i=1}^{n} l_{ik}.
\label{eq:vote_C1}
\end{equation}
If $\mathbf{u}_k$ is preferred over many anchors, then many entries satisfy $y_{ik}=-1$, leading to a larger $\mathrm{score}(\mathbf{u}_k)$. Hence, higher scores indicate better candidates.

\paragraph{C2 voting}
For the C2 criterion, $\mathbf{L}\in\{-1,0,+1\}^{n\times q}$. In addition, each anchor $\mathbf{x}_i$ is assigned a category label during data preparation (e.g., good vs.\ bad). Let $\mathcal{I}$ and $\mathcal{II}$ denote the index sets of good and bad anchors, respectively, with $|\mathcal{I}|+|\mathcal{II}|=n$.

For a candidate $\mathbf{u}_k$, we count the relation outcomes within each group:
\begin{equation}
c_k^{t}(l)=\left|\left\{ i\in t \;:\; l_{ik} \right\}\right|,
t\in\{\mathcal{I},\mathcal{II}\},\ l\in\{-1,0,+1\}.
\end{equation}
That is, $c_k^{\mathcal{I}}(+1)$ counts how many good anchors judge the candidate as worse ($l_{ik}=+1$), while $c_k^{\mathcal{I}}(-1)$ counts how many judge it as better ($l_{ik}=-1)$; analogous definitions hold for the bad-anchor group $\mathcal{II}$.

We then compute the normalized voting score:
\begin{equation}
\begin{split}
\mathrm{score}_{C2}(\mathbf{u}_k)=\frac{1}{n}\big(
& c_k^{\mathcal{I}}(0) + c_k^{\mathcal{I}}(-1) + c_k^{\mathcal{II}}(-1) \\
& - c_k^{\mathcal{I}}(+1) - c_k^{\mathcal{II}}(+1) - c_k^{\mathcal{II}}(0)
\big).
\end{split}
\label{eq:vote_C2}
\end{equation}
Since the score is averaged by $n$, $\mathrm{score}(\mathbf{u}_k)\in[-1,1]$. A positive value indicates that $\mathbf{u}_k$ is more likely to belong to the good category, while a negative value suggests the opposite; the magnitude reflects confidence.

In summary, the above pipeline, including data preparation, anchor-based context engineering, and voting-based aggregation, enables relation prediction and ranking using a general LLM. However, the difficulty of relational reasoning and the latency and cost of frontier LLMs motivate us to further specialize compact LLMs for this task via reinforcement learning, as described next.

\subsection{Fine-tuning via Reinforcement Learning}
\label{subsec:proposed_method_RL}
While the pipeline in Section~\ref{subsec:proposed_method_TD} enables relation prediction with a general LLM, using frontier models inside an EA loop is often too costly, and smaller models may lack sufficient relational reasoning ability. We therefore fine-tune Qwen2.5-7B-Instruct with Group Relative Policy Optimization (GRPO)~\cite{shao2024deepseekmath}, which avoids a separate critic and readily supports rule-based rewards for discrete relation labels and output formats.

\subsubsection{GRPO Objective in Our Task}
\label{subsec:grpo_in_task}

Let $\pi_{\theta}$ denote the policy (LLM) to be optimized and $\pi_{\mathrm{ref}}$ the frozen reference policy (the base model) used for KL regularization. 
For each prompt $\mathbf{s}$ (built from one anchor and its associated in-context examples and query pairs), we sample a \emph{group} of $G$ responses
$\{\mathbf{y}^{(g)}\}_{g=1}^{G}\sim \pi_{\theta}(\cdot\mid \mathbf{s})$.
Each response $\mathbf{y}^{(g)}$ is a JSON object containing predicted relation labels for all $q$ query pairs in $\mathbf{s}$ (Section~\ref{subsec:AnchorBasedEngineering}). 
We evaluate each response with a scalar reward
$r^{(g)} = R(\mathbf{y}^{(g)},\mathbf{y}^{*})$,
where $\mathbf{y}^{*}$ is the ground-truth JSON and $R(\cdot)$ is defined in Section~\ref{subsec:reward_design}.

GRPO updates $\pi_{\theta}$ by increasing the likelihood of sampled responses with higher reward (relative to other responses for the same prompt), while preventing the policy from drifting too far from $\pi_{\mathrm{ref}}$. 
In our implementation, we use the following compact GRPO objective:
\begin{equation}
\label{eq:grpo_obj_simplified}
\begin{aligned}
\max_{\theta}\;\mathbb{E}_{\mathbf{s}}
\Bigg[
\frac{1}{G}\sum_{g=1}^{G}
\Bigg(
&\min\Big(
\rho^{(g)}\hat{A}^{(g)}, \\
&\quad \mathrm{clip}\!\big(\rho^{(g)},1-\epsilon,1+\epsilon\big)\hat{A}^{(g)}
\Big) \\
&\quad -\beta\,\mathbb{D}_{\mathrm{KL}}\!\big(\pi_{\theta}\,\|\,\pi_{\mathrm{ref}}\big)
\Bigg)
\Bigg].
\end{aligned}
\end{equation}
where $\epsilon$ is the clipping threshold and $\beta$ controls the KL penalty. 
Here, the probability ratio is
\begin{equation}
\label{eq:grpo_ratio_simple}
\rho^{(g)}=
\frac{\pi_{\theta}\!\left(\mathbf{y}^{(g)}\mid \mathbf{s}\right)}
{\pi_{\theta_{\mathrm{old}}}\!\left(\mathbf{y}^{(g)}\mid \mathbf{s}\right)},
\end{equation}
and $\pi_{\theta_{\mathrm{old}}}$ is the snapshot of the policy $\pi_\theta$ at the beginning of the optimization epoch, which acts as the behavior policy used to sample $\mathbf{y}^{(g)}$.

Unlike PPO, GRPO does not learn a value function. Instead, it computes the advantage within the group by normalizing rewards across the $G$ sampled responses for the same prompt:
\begin{equation}
\label{eq:grpo_adv}
\hat{A}^{(g)}=
\frac{r^{(g)}-\mathrm{mean}(\mathbf{r})}{\mathrm{std}(\mathbf{r})+\delta},
\quad 
\mathbf{r}=\big[r^{(1)},\ldots,r^{(G)}\big],
\end{equation}
where $\delta$ is a small constant. 
In our relational setting, Eq.~\eqref{eq:grpo_adv} encourages the model to prefer, for the same anchor-based prompt, those JSON predictions that achieve higher relation accuracy and valid formatting relative to other sampled outputs, thereby providing a stable, low-variance learning signal tailored to relation reasoning.

\subsubsection{Training Data Construction and Preprocessing}
\label{subsec:train_data}
The RL training corpus is built from evolutionary trajectories on a suite of benchmark problems, including LZG~\cite{liu2013gaussian}, CEC2008~\cite{tang2007benchmark}, and CEC2010~\cite{mallipeddi2010problem}. For each function, we run a genetic algorithm (GA) for 100 generations and record population snapshots every 10 generations to capture diverse optimization states.
During GA execution, both parent and offspring populations are maintained at $N=100$. For dataset construction, we downsample each snapshot to 30 individuals, decoupling the execution size from the observation size. This design serves two purposes:
\begin{enumerate}
    \item Inference efficiency: using 30 individuals reduces prompt length and accelerates LLM inference/training.
    \item Diversity preservation: maintaining $N=100$ during GA helps preserve search diversity and avoids premature convergence~\cite{chen2012large}; downsampling from a larger pool yields more varied contexts for the model to learn from~\cite{wang2022evolutionary,schonemann2004impact}.
\end{enumerate}

Each snapshot is normalized independently by min--max scaling of decision variables to $[0,1]$. Using the prompt construction described in Section~\ref{subsec:AnchorBasedEngineering} and the relation definitions in C1/C2, we generate 8640 instruction pairs. Each response provides the ground-truth relation labels (mapped to $\{+1,-1\}$ for C1 and $\{+1,0,-1\}$ for C2), which are then used to compute rewards in RL.

\subsubsection{Reward Function Design}
\label{subsec:reward_design}
To align GRPO with our downstream use case (LLM assisted selection), we design a unified rule-based reward $R(\mathbf{y},\mathbf{y}^{*})$ that evaluates a sampled response $\mathbf{y}$ against the ground truth $\mathbf{y}^{*}$ along three dimensions:
\begin{enumerate}
    \item Format compliance: the output must be a valid JSON object with exactly $q$ predicted labels, matching the $q$ candidates in $\mathcal{D}_{\mathrm{query}}$ for that prompt. Any violation receives a fixed penalty.
    \item Relation accuracy: if the format is valid, we reward the fraction of correctly predicted labels, which directly corresponds to the correctness of pairwise relations used later for voting-based scoring.
    \item Output diversity: to reduce entropy collapse (e.g., predicting a single label for all candidates), we discount the reward if one label dominates the output.
\end{enumerate}
Formally, let $\mathcal{V}(\mathbf{y})$ be a validation function for JSON format and item count, $N_{\mathrm{correct}}$ be the number of correctly predicted labels, and $q$ be the number of candidates in the query set. The reward is defined as
\begin{equation}
\label{eq:reward}
R(\mathbf{y},\mathbf{y}^{*})=
\begin{cases}
-0.2, & \text{if } \mathcal{V}(\mathbf{y})=\mathrm{False},\\[3pt]
\lambda \cdot \dfrac{N_{\mathrm{correct}}}{q}, & \text{if } \mathcal{V}(\mathbf{y})=\mathrm{True},
\end{cases}
\end{equation}
where the diversity coefficient $\lambda$ is
\begin{equation}
\label{eq:lambda}
\lambda=
\begin{cases}
0.8, & \text{if } \max\limits_{l\in \mathcal{L}}\left(\dfrac{\mathrm{count}(l)}{q}\right)>\tau,\\[6pt]
1.0, & \text{otherwise},
\end{cases}
\end{equation}
$\mathcal{L}$ is the set of labels appearing in $\mathbf{y}$, and $\tau=0.9$ in our experiments. This reward design provides dense feedback (partial credit) while discouraging trivial degenerate outputs, thus better matching the needs of selection in surrogate-assisted EAs.

\subsubsection{Training Settings}
\label{subsec:training_settings}
The RL fine-tuning was conducted on 16 NVIDIA H100 GPUs. We trained for 10 epochs with learning rate $1\times10^{-6}$. We set the GRPO group size to $G=8$, i.e., the model generates 8 candidate JSON responses per prompt to compute the group-normalized advantages in Eq.~\eqref{eq:grpo_adv}.

Based on this setup, the trained model is referred to as ReLLM. The models trained under criteria C1 and C2 are denoted as ReLLM-C1 and ReLLM-C2, respectively.

\subsection{LLM-Assisted Evolutionary Algorithm}
\label{subsec:proposed_method_LLM4EA}
To evaluate the proposed relational LLM surrogate in practical optimization, we integrate it into a standard SAEA framework~\cite{jin2011surrogateassisted}. Unlike conventional SAEAs that retrain a surrogate each generation, our method uses an inference-only LLM: anchor-based prompts infer pairwise relations, which are then converted into absolute scores via voting. Therefore, in each generation the surrogate step mainly involves constructing the in-context set from the archive, rather than gradient-based retraining.

Throughout the evolutionary process, all truly evaluated solutions are stored in an archive $\mathcal{A}$, which serves as the knowledge base for prompt construction. In each generation, after generating an offspring set $\mathcal{Q}$, we sample (or truncate) at most $\tau$ solutions from $\mathcal{A}$ to form the context set $\mathcal{D}_{\mathrm{ctx}}$ for the LLM surrogate; the candidates to be screened correspond to $\mathcal{D}_{\mathrm{query}}=\mathcal{Q}$. 

Given $\mathcal{D}_{\mathrm{ctx}}$ and $\mathcal{D}_{\mathrm{query}}$, the LLM predicts the relation matrix $\mathbf{L}$ (under C1 or C2), which is then aggregated into a scalar $\mathrm{score}(\mathbf{u}_k)$ for each candidate via voting. The algorithm selects $N'$ promising solutions $\mathcal{Q}'\subset\mathcal{Q}$ for expensive evaluation, thereby reducing the number of real evaluations per generation under tight budgets. The overall procedure is summarized in Algorithm~\ref{alg:sas}.

\begin{algorithm}[ht!]
\caption{Framework of the R2SAEA}
\label{alg:sas}
\SetKwInOut{Input}{Input}\SetKwInOut{Output}{Output}
\Input{$\mathbf{Gen}$ (variation operator), $\mathbf{Sel}$ (environmental selection), $N$ (population size), $N'$ (evaluation size), $\tau$ (max context size)}
\Output{$s$ (single optimum) / $S$ (set of non-dominated solutions)}
\BlankLine
$\mathcal{A}, \mathcal{P} \leftarrow \mathrm{Initialize}(N)$\; \label{alg:sas:init}
\While{not $\mathrm{StoppingCondition()}$}{
    $\mathcal{Q} \leftarrow \mathbf{Gen}(\mathcal{P}, N)$\; \label{alg:sas:gen}
    $\mathcal{D}_{\mathrm{ctx}} \leftarrow \mathrm{SampleContext}(\mathcal{A}, \tau)$\; \label{alg:sas:ctx}
    $\mathcal{D}_{\mathrm{query}} \leftarrow \mathcal{Q}$\; \label{alg:sas:query}
    $\mathbf{L} \leftarrow \mathrm{LLMRelationInfer}(\mathcal{D}_{\mathrm{ctx}}, \mathcal{D}_{\mathrm{query}})$\; \label{alg:sas:infer}
    $\mathbf{s} \leftarrow \mathrm{VoteAggregate}(\mathbf{L})$\; \label{alg:sas:vote}
    $\mathcal{Q}' \leftarrow \mathrm{SelectTop}(\mathcal{D}_{\mathrm{query}}, \mathbf{s}, N')$\; \label{alg:sas:sasel}
    $\mathcal{Q}' \leftarrow \mathrm{Evaluate}(\mathcal{Q}')$\; \label{alg:sas:eval}
    $\mathcal{P} \leftarrow \mathbf{Sel}(\mathcal{P}\cup \mathcal{Q}', N)$\; \label{alg:sas:sel}
    $\mathcal{A} \leftarrow \mathcal{A}\cup \mathcal{Q}'$\; \label{alg:sas:archive}
}
\end{algorithm}

This framework applies to both single- and multi-objective optimization; in this paper, the resulting algorithms are denoted as R2SAEA-SOP and R2SAEA-MOP. The main differences lie in the relation criterion (C1/C2) and the underlying EA operators (e.g., $\mathbf{Sel}$ and performance indicators). Concrete instantiations are provided in the supplementary material.

\section{Empirical Studies}
\label{sec:experiments}
This section evaluates the proposed relational LLM surrogate for expensive optimization. We first integrate it into single-objective and multi-objective SAEAs and conduct statistical comparisons against state-of-the-art baselines under limited evaluation budgets. We then use a sampled offline dataset to compare relation prediction accuracy with representative general LLMs. Next, we study robustness to data scale, including decision-space dimensionality and context/query size. Finally, we analyze the effects of normalization strategies and model quantization on accuracy and deployment efficiency.

\subsection{Empirical Studies in SOPs}
\label{subsec:experiments_sops}

\subsubsection{Algorithms for comparison}

We evaluate the proposed single-objective framework (denoted as \textbf{R2SAEA-SOP}) against representative baselines covering: (1) an EA without surrogates, (2) an LLM-based surrogate approach that relies on general-purpose models, (3) Bayesian optimization, and (iv) classical SAEAs based on regression, classification, and relation surrogates. The compared algorithms are summarized as follows.
\begin{itemize}
    \item \textbf{Baseline EA:} We adopt EDA/LS~\cite{zhou2015estimation} as the baseline optimizer. This choice ensures a fair comparison in terms of the search operator, since EDA/LS, LSEA, and ReLLM-S all employ the Variable Width Histogram (VWH) model for reproduction.
    
    \item \textbf{General LLM as surrogate:} We include LSEA~\cite{hao2024large}, which uses prompted general-purpose LLMs as regression and/or classification surrogates. Following its original setting, we consider four variants:
    \begin{itemize}
        \item LAEA-8B: Llama3-8B as both regression and classification surrogates.
        \item LAEA-Reg-8B: Llama3-8B as regression surrogate only.
        \item LAEA-8x7B: Mixtral-8$\times$7B as both regression and classification surrogates.
        \item LAEA-Reg-8x7B: Mixtral-8$\times$7B as regression surrogate only.
    \end{itemize}
    These baselines reflect the current practice of using general LLMs via prompting, and thus directly test whether our RL-specialized relational LLM surrogate is beneficial.
    \item \textbf{Bayesian optimization:} We include Bayesian optimization (BO)~\cite{stander2002robustness} as a strong sample-efficient baseline widely used in expensive black-box optimization.
    \item \textbf{Conventional SAEAs:} We compare with several SOTA SAEAs implemented in PlatEMO~\cite{tian2017platemo}, including SACOSO~\cite{sun2017surrogate} and EGO~\cite{jones1998efficient} (regression-based), SAMSO~\cite{li2020surrogate} (surrogate-assisted memetic search), and DRSO~\cite{hao2024enhancing} (relation-based). We further include FCPS~\cite{zhou2019fuzzy}, a classifier-assisted differential evolution algorithm.
\end{itemize}

To ensure fair comparisons, we use the recommended parameter settings reported in the original literature for each baseline\footnote{BO is executed with default package settings; SAMSO and other PlatEMO algorithms follow the PlatEMO default configurations~\cite{tian2017platemo}; FCPS and EDA/LS are implemented by us according to their original descriptions.}. The main settings are:
\begin{itemize}
  \item Evaluation budget: $FEs = 300$.
  \item Population size: The default setting is $N=30$, SACOSO, EGO, SAMSO are set to $N=50$.
  \item Dimensionality: $D\in\{5,10,20\}$.
\end{itemize}

Each algorithm is independently run 30 times on each instance. Statistical significance is assessed using the Wilcoxon rank-sum test~\cite{hollander2013nonparametric} at the 0.05 level. In result tables, symbols `+', `-', and `$\approx$' indicate that a comparison algorithm is significantly better than, worse than, or statistically similar to R2SAEA-SOP, respectively.

\subsubsection{Benchmark Problems}
We use two standard expensive single-objective benchmark suites: LZG~\cite{liu2013gaussian} and YLL~\cite{yao1999evolutionary}. The LZG suite contains four classical functions (Ellipsoid, Rosenbrock, Ackley, and Griewank), covering unimodal convex landscapes, narrow valley structures, and highly multimodal terrains. The YLL suite (YLLF01--YLLF13) includes diverse characteristics such as multimodality, stepwise behaviors, and noise; YLLF10 and YLLF11 are excluded due to overlap with LZG. Since LZG is also used in later analyses (e.g., sensitivity studies), we briefly summarize its properties here: Ellipsoid is strictly unimodal; Rosenbrock exhibits a long narrow valley; Ackley and Griewank are multimodal with many local optima.

\subsubsection{Empirical Results}
\begin{figure}[ht]
    \centering
    \includegraphics[width=1\linewidth]{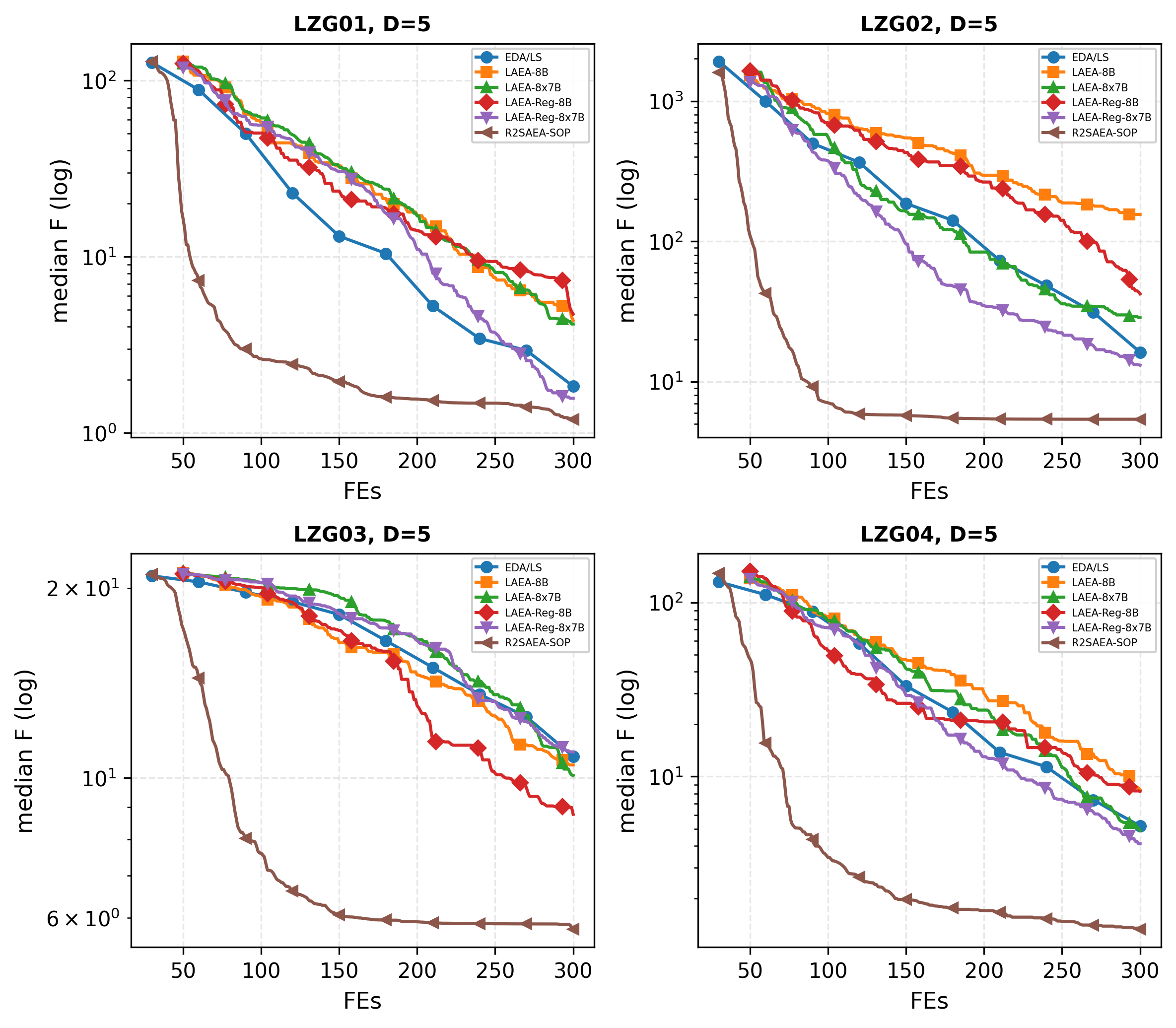}
    \caption{Median runtime performance of LSEA and R2SAEA-SOP on the LZG test suite over 30 independent runs, with dimension $D=5$.}
    \label{fig:lzg_runtime_5d}
\end{figure}
\definecolor{hl}{rgb}{0.75,0.75,0.75}
\sethlcolor{hl}

\begin{table*}[ht!]
\renewcommand{\arraystretch}{1.2}
\centering
\caption{Statistics of mean and standard deviation results obtained by six comparison algorithms on LZG and YLL test suites with $D=5,10,20$.}
\resizebox{\textwidth}{!}{%
\begin{tabular}{ccccccccc}
\toprule
Problem & $D$ & BayesianOptimization & SACOSO & DRSO & EGO & FCPS & SAMSO & R2SAEA-SOP \\
\midrule
\multirow{3}{*}{LZG01} & 5 & \hl{2.4729e-03(5.53e-04)[1] $+$} & 1.3277e+00(1.62e+00)[7] $-$ & 7.2798e-02(2.48e-01)[3] $+$ & 5.0789e-03(7.16e-03)[2] $+$ & 2.5047e-01(2.81e-01)[5] $\approx$ & 9.9509e-02(1.33e-01)[4] $+$ & 3.4616e-01(4.59e-01)[6] \\
 & 10 & 3.7479e-02(5.84e-03)[2] $+$ & 1.7285e+01(1.67e+01)[6] $-$ & 2.0871e+00(4.79e+00)[3] $+$ & \hl{2.5678e-02(1.92e-02)[1] $+$} & 1.7713e+01(9.68e+00)[7] $-$ & 6.5646e+00(7.32e+00)[5] $-$ & 2.7559e+00(2.78e+00)[4] \\
 & 20 & \hl{3.3135e-01(6.98e-02)[1] $+$} & 1.8407e+02(1.28e+02)[5] $-$ & 3.9882e+01(3.77e+01)[4] $-$ & 4.3635e+02(4.49e+02)[7] $\approx$ & 1.9926e+02(5.54e+01)[6] $-$ & 2.5199e+01(2.41e+01)[3] $\approx$ & 1.7799e+01(7.01e+00)[2] \\
\hline
\multirow{3}{*}{LZG02} & 5 & 1.6513e+00(1.09e+00)[2] $+$ & 2.0388e+01(9.75e+00)[7] $-$ & 7.4108e+00(1.11e+01)[6] $\approx$ & \hl{3.3821e-01(3.09e-01)[1] $+$} & 5.6777e+00(2.34e+00)[4] $\approx$ & 5.6327e+00(3.43e+00)[3] $\approx$ & 5.6816e+00(3.15e+00)[5] \\
 & 10 & 3.8266e+01(9.95e+00)[3] $-$ & 1.0775e+02(8.06e+01)[7] $-$ & 5.3948e+01(3.20e+01)[5] $-$ & 3.8817e+01(6.47e+01)[4] $\approx$ & 8.1967e+01(3.05e+01)[6] $-$ & 3.6444e+01(1.48e+01)[2] $-$ & \hl{2.3680e+01(1.08e+01)[1]} \\
 & 20 & 1.9292e+02(5.78e+01)[4] $-$ & 6.1024e+02(4.51e+02)[6] $-$ & 1.9026e+02(1.00e+02)[3] $-$ & 1.6121e+03(1.43e+03)[7] $-$ & 5.2840e+02(1.42e+02)[5] $-$ & \hl{5.1047e+01(2.93e+01)[1] $\approx$} & 5.4580e+01(1.41e+01)[2] \\
\hline
\multirow{3}{*}{LZG03} & 5 & \hl{2.3538e+00(6.13e-01)[1] $+$} & 5.5709e+00(1.91e+00)[6] $-$ & 4.1793e+00(3.38e+00)[5] $\approx$ & 4.0706e+00(4.09e+00)[4] $\approx$ & 3.7299e+00(2.46e+00)[2] $\approx$ & 1.2303e+01(3.51e+00)[7] $-$ & 3.9031e+00(2.14e+00)[3] \\
 & 10 & \hl{3.8021e+00(1.37e+00)[1] $+$} & 6.4929e+00(1.96e+00)[4] $\approx$ & 4.7955e+00(2.34e+00)[2] $\approx$ & 1.7842e+01(2.10e+00)[7] $-$ & 1.1965e+01(1.66e+00)[5] $-$ & 1.7664e+01(1.81e+00)[6] $-$ & 5.4982e+00(1.47e+00)[3] \\
 & 20 & \hl{6.8505e+00(2.09e+00)[1] $+$} & 7.5181e+00(3.29e+00)[2] $+$ & 1.0235e+01(2.02e+00)[4] $-$ & 1.9444e+01(5.50e-01)[7] $-$ & 1.6160e+01(1.29e+00)[5] $-$ & 1.8702e+01(1.30e+00)[6] $-$ & 8.1405e+00(1.11e+00)[3] \\
\hline
\multirow{3}{*}{LZG04} & 5 & 5.7971e-01(1.40e-01)[2] $\approx$ & 1.5276e+00(1.19e+00)[6] $-$ & \hl{3.8259e-01(4.27e-01)[1] $+$} & 8.7939e-01(2.02e-01)[3] $\approx$ & 1.4571e+00(1.03e+00)[5] $\approx$ & 1.9405e+00(1.22e+00)[7] $-$ & 1.4313e+00(1.64e+00)[4] \\
 & 10 & \hl{1.0609e+00(9.70e-02)[1] $+$} & 5.9792e+00(4.83e+00)[5] $-$ & 1.9643e+00(1.63e+00)[2] $+$ & 4.4939e+00(1.87e+01)[4] $-$ & 1.3391e+01(7.24e+00)[6] $-$ & 1.5778e+01(1.42e+01)[7] $-$ & 2.2656e+00(1.10e+00)[3] \\
 & 20 & \hl{1.4509e+00(1.66e-01)[1] $+$} & 4.7279e+01(3.48e+01)[5] $-$ & 1.3291e+01(7.67e+00)[3] $-$ & 2.3290e+02(1.51e+02)[7] $-$ & 8.7833e+01(2.08e+01)[6] $-$ & 2.6314e+01(1.77e+01)[4] $-$ & 7.2017e+00(2.50e+00)[2] \\
\hline
\multirow{3}{*}{YLLF01} & 5 & \hl{1.6692e-01(7.14e-02)[1] $+$} & 1.4218e+02(1.98e+02)[7] $-$ & 2.0378e-01(6.48e-01)[2] $+$ & 7.8895e-01(9.59e-01)[3] $+$ & 6.2088e+01(7.91e+01)[5] $\approx$ & 7.2620e+01(9.81e+01)[6] $\approx$ & 3.9479e+01(6.93e+01)[4] \\
 & 10 & \hl{2.0593e+00(6.54e-01)[1] $+$} & 1.4075e+03(1.76e+03)[6] $-$ & 9.7451e+01(2.70e+02)[2] $+$ & 7.3814e+02(2.76e+03)[4] $-$ & 1.4524e+03(6.27e+02)[7] $-$ & 1.0534e+03(9.53e+02)[5] $-$ & 1.3397e+02(9.93e+01)[3] \\
 & 20 & \hl{1.3129e+01(3.59e+00)[1] $+$} & 7.0639e+03(4.95e+03)[5] $-$ & 1.7870e+03(2.02e+03)[4] $-$ & 2.4236e+04(1.67e+04)[7] $-$ & 8.2352e+03(2.25e+03)[6] $-$ & 1.3785e+03(1.90e+03)[3] $\approx$ & 7.0346e+02(2.45e+02)[2] \\
\hline
\multirow{3}{*}{YLLF02} & 5 & 9.0472e+00(3.66e+00)[6] $-$ & 3.1914e+00(1.64e+00)[4] $-$ & \hl{2.7634e-02(6.98e-02)[1] $+$} & 1.0713e+01(3.05e+00)[7] $-$ & 8.4681e-01(5.90e-01)[3] $\approx$ & 5.0546e+00(2.34e+00)[5] $-$ & 6.6314e-01(4.67e-01)[2] \\
 & 10 & 8.7108e+01(2.14e+02)[6] $-$ & 1.0470e+01(5.30e+00)[4] $-$ & \hl{5.9244e-01(8.13e-01)[1] $+$} & 2.0656e+02(3.18e+02)[7] $-$ & 7.5368e+00(2.35e+00)[3] $-$ & 2.1132e+01(1.48e+01)[5] $-$ & 2.0483e+00(9.42e-01)[2] \\
 & 20 & 5.8266e+04(1.57e+05)[6] $-$ & 8.3429e+02(4.26e+03)[4] $-$ & 1.1547e+01(3.66e+00)[2] $\approx$ & 6.0469e+06(9.18e+06)[7] $-$ & 2.7616e+01(3.48e+00)[3] $-$ & 8.0920e+03(4.32e+04)[5] $-$ & \hl{1.0415e+01(2.44e+00)[1]} \\
\hline
\multirow{3}{*}{YLLF03} & 5 & \hl{2.3036e+00(1.25e+00)[1] $+$} & 6.4408e+02(5.16e+02)[7] $-$ & 4.2269e+02(4.02e+02)[6] $-$ & 1.5975e+01(4.39e+01)[2] $+$ & 8.9536e+01(1.19e+02)[3] $\approx$ & 1.7554e+02(3.34e+02)[5] $\approx$ & 1.3389e+02(2.42e+02)[4] \\
 & 10 & 9.5773e+03(1.86e+03)[7] $-$ & 5.6777e+03(2.21e+03)[6] $-$ & 2.9936e+03(1.42e+03)[4] $-$ & 3.7617e+02(1.35e+03)[2] $-$ & 3.4386e+03(1.19e+03)[5] $-$ & 2.6782e+03(2.79e+03)[3] $-$ & \hl{2.9312e+02(1.78e+02)[1]} \\
 & 20 & 3.7928e+04(8.01e+03)[6] $-$ & 3.4882e+04(9.56e+03)[5] $-$ & 2.0508e+04(6.40e+03)[4] $-$ & 4.5902e+04(1.46e+04)[7] $-$ & 1.7537e+04(4.61e+03)[2] $-$ & 2.0077e+04(1.48e+04)[3] $-$ & \hl{7.5158e+03(3.12e+03)[1]} \\
\hline
\multirow{3}{*}{YLLF04} & 5 & \hl{9.0981e-01(4.21e-01)[1] $+$} & 7.7770e+00(3.55e+00)[5] $\approx$ & 1.6121e+01(1.27e+01)[7] $-$ & 2.2182e+00(4.09e+00)[2] $+$ & 5.7313e+00(4.12e+00)[3] $\approx$ & 1.5364e+01(6.74e+00)[6] $-$ & 7.2501e+00(5.61e+00)[4] \\
 & 10 & \hl{5.4310e+00(2.31e+00)[1] $+$} & 1.6539e+01(6.60e+00)[3] $-$ & 3.4108e+01(1.48e+01)[6] $-$ & 2.4192e+01(1.65e+01)[4] $-$ & 2.8546e+01(6.82e+00)[5] $-$ & 4.7709e+01(9.76e+00)[7] $-$ & 1.0091e+01(5.67e+00)[2] \\
 & 20 & 3.9170e+01(1.17e+01)[3] $-$ & 3.3475e+01(1.42e+01)[2] $-$ & 5.0072e+01(1.41e+01)[4] $-$ & 6.4871e+01(6.98e+00)[6] $-$ & 5.2011e+01(5.70e+00)[5] $-$ & 7.3798e+01(6.56e+00)[7] $-$ & \hl{1.4952e+01(2.62e+00)[1]} \\
\hline
\multirow{3}{*}{YLLF05} & 5 & 3.6052e+04(2.73e+04)[7] $-$ & 1.8244e+04(3.59e+04)[6] $-$ & \hl{5.5278e+02(9.20e+02)[1] $\approx$} & 5.6275e+02(5.56e+02)[2] $+$ & 1.4976e+03(1.65e+03)[3] $\approx$ & 2.9979e+03(3.74e+03)[5] $-$ & 2.5850e+03(8.09e+03)[4] \\
 & 10 & 3.6812e+05(1.90e+05)[4] $-$ & 7.2103e+05(1.21e+06)[6] $-$ & 4.5238e+04(1.27e+05)[2] $-$ & 2.5608e+06(8.79e+06)[7] $-$ & 6.3924e+05(5.55e+05)[5] $-$ & 1.1711e+05(1.24e+05)[3] $-$ & \hl{5.7033e+03(9.24e+03)[1]} \\
 & 20 & 3.5162e+06(1.26e+06)[4] $-$ & 4.6213e+06(6.80e+06)[5] $-$ & 2.2437e+06(4.88e+06)[3] $-$ & 9.8184e+07(4.29e+07)[7] $-$ & 1.0486e+07(6.08e+06)[6] $-$ & 1.8144e+05(1.10e+05)[2] $-$ & \hl{8.5401e+04(6.02e+04)[1]} \\
\hline
\multirow{3}{*}{YLLF06} & 5 & \hl{0.0000e+00(0.00e+00)[1] $+$} & 1.9140e+02(3.84e+02)[7] $\approx$ & 5.1500e+01(1.21e+02)[4] $\approx$ & 1.8000e+00(1.56e+00)[2] $+$ & 6.8733e+01(6.24e+01)[6] $-$ & 5.7200e+01(6.71e+01)[5] $-$ & 4.1840e+01(7.55e+01)[3] \\
 & 10 & \hl{2.5667e+00(8.83e-01)[1] $+$} & 8.3087e+02(9.64e+02)[6] $-$ & 2.5653e+02(6.10e+02)[3] $\approx$ & 5.1930e+02(2.21e+03)[4] $-$ & 1.5938e+03(7.35e+02)[7] $-$ & 7.7283e+02(1.09e+03)[5] $-$ & 9.4960e+01(8.57e+01)[2] \\
 & 20 & \hl{1.2767e+01(3.69e+00)[1] $+$} & 7.6831e+03(5.40e+03)[5] $-$ & 1.4370e+03(8.29e+02)[3] $-$ & 2.4539e+04(1.76e+04)[7] $-$ & 8.2558e+03(2.42e+03)[6] $-$ & 2.2655e+03(2.38e+03)[4] $-$ & 7.7716e+02(3.67e+02)[2] \\
\hline
\multirow{3}{*}{YLLF07} & 5 & 9.9383e-02(4.93e-02)[5] $-$ & 1.2533e-01(9.35e-02)[6] $-$ & 6.0171e-02(5.44e-02)[3] $-$ & 8.7505e-02(4.15e-02)[4] $-$ & 5.5496e-02(3.62e-02)[2] $-$ & 1.3369e-01(8.83e-02)[7] $-$ & \hl{1.5794e-02(1.09e-02)[1]} \\
 & 10 & 3.4966e-01(1.42e-01)[4] $-$ & 4.5793e-01(3.34e-01)[6] $-$ & 4.4162e-01(6.71e-01)[5] $-$ & 9.7304e-01(1.66e+00)[7] $-$ & 3.4575e-01(1.58e-01)[3] $-$ & 2.5794e-01(1.51e-01)[2] $-$ & \hl{4.7159e-02(2.50e-02)[1]} \\
 & 20 & 1.4840e+00(4.79e-01)[4] $-$ & 3.6796e+00(3.93e+00)[5] $-$ & 5.2634e-01(3.16e-01)[3] $-$ & 3.6703e+01(1.30e+01)[7] $-$ & 3.9538e+00(1.21e+00)[6] $-$ & 3.9917e-01(1.85e-01)[2] $-$ & \hl{1.5448e-01(6.19e-02)[1]} \\
\hline
\multirow{3}{*}{YLLF08} & 5 & 4.2772e+02(2.17e+02)[3] $+$ & 1.0046e+03(1.48e+02)[7] $\approx$ & \hl{1.8297e+02(1.42e+02)[1] $+$} & 3.7095e+02(1.37e+02)[2] $+$ & 8.3052e+02(1.13e+02)[4] $+$ & 9.5805e+02(1.13e+02)[6] $\approx$ & 9.3211e+02(1.78e+02)[5] \\
 & 10 & 1.9175e+03(5.07e+02)[3] $+$ & 2.6298e+03(2.45e+02)[7] $-$ & \hl{5.1615e+02(2.24e+02)[1] $+$} & 1.2103e+03(3.13e+02)[2] $+$ & 2.4063e+03(1.77e+02)[4] $+$ & 2.4090e+03(2.79e+02)[5] $\approx$ & 2.4883e+03(2.54e+02)[6] \\
 & 20 & 5.8570e+03(3.89e+02)[5] $\approx$ & 5.8486e+03(3.57e+02)[4] $\approx$ & \hl{2.9454e+03(6.45e+02)[1] $+$} & 3.8476e+03(6.13e+02)[2] $+$ & 5.7475e+03(3.46e+02)[3] $\approx$ & 5.8709e+03(3.56e+02)[6] $\approx$ & 5.8794e+03(2.46e+02)[7] \\
\hline
\multirow{3}{*}{YLLF09} & 5 & 1.2630e+01(5.24e+00)[5] $-$ & 2.2590e+01(9.21e+00)[7] $-$ & 5.1785e+00(2.65e+00)[3] $-$ & 6.5478e+00(3.18e+00)[4] $-$ & \hl{2.5262e+00(2.81e+00)[1] $\approx$} & 1.5650e+01(8.35e+00)[6] $-$ & 3.2886e+00(1.87e+00)[2] \\
 & 10 & 4.4701e+01(2.31e+01)[5] $-$ & 6.7066e+01(1.71e+01)[7] $-$ & \hl{1.8525e+01(7.22e+00)[1] $+$} & 3.9948e+01(1.76e+01)[4] $-$ & 2.8772e+01(1.13e+01)[3] $\approx$ & 6.3717e+01(2.21e+01)[6] $-$ & 2.7442e+01(1.00e+01)[2] \\
 & 20 & 1.8391e+02(2.32e+01)[6] $-$ & 1.7510e+02(3.86e+01)[5] $-$ & \hl{1.1107e+02(2.15e+01)[1] $+$} & 2.0057e+02(3.70e+01)[7] $-$ & 1.1497e+02(1.86e+01)[2] $+$ & 1.2867e+02(4.79e+01)[3] $\approx$ & 1.3046e+02(1.61e+01)[4] \\
\hline
\multirow{3}{*}{YLLF12} & 5 & 4.5591e+03(1.04e+04)[7] $-$ & 4.0684e+02(2.13e+03)[5] $-$ & 6.5408e+01(2.48e+02)[4] $\approx$ & 8.4486e+02(2.77e+03)[6] $-$ & 7.0874e+00(4.86e+00)[2] $-$ & 1.1565e+01(1.02e+01)[3] $-$ & \hl{3.8463e+00(5.12e+00)[1]} \\
 & 10 & 8.7095e+05(9.79e+05)[6] $-$ & 9.3171e+04(4.31e+05)[4] $-$ & 3.2607e+04(1.73e+05)[3] $-$ & 5.3121e+06(1.10e+07)[7] $-$ & 1.1621e+05(1.85e+05)[5] $-$ & 6.3152e+02(2.10e+03)[2] $-$ & \hl{7.6018e+00(1.34e+01)[1]} \\
 & 20 & 1.0138e+07(7.51e+06)[5] $-$ & 5.5863e+06(9.48e+06)[4] $-$ & 8.3288e+05(2.14e+06)[3] $-$ & 2.2320e+08(8.82e+07)[7] $-$ & 1.0731e+07(1.17e+07)[6] $-$ & 7.8787e+02(2.12e+03)[2] $-$ & \hl{1.4365e+01(6.86e+00)[1]} \\
\hline
\multirow{3}{*}{YLLF13} & 5 & 1.2911e+10(1.01e+10)[7] $-$ & 2.4962e+09(5.86e+09)[6] $-$ & 4.8965e+07(2.51e+08)[2] $+$ & 3.2823e+08(2.44e+08)[3] $+$ & \hl{1.3471e+02(5.05e+02)[1] $+$} & 1.6399e+09(2.78e+09)[5] $-$ & 8.7035e+08(3.45e+09)[4] \\
 & 10 & 2.3671e+11(1.36e+11)[6] $-$ & 6.6608e+10(1.79e+11)[5] $-$ & 1.3545e+09(2.33e+09)[2] $\approx$ & 3.2521e+11(9.28e+11)[7] $-$ & \hl{1.9126e+06(2.36e+06)[1] $+$} & 1.9201e+10(2.33e+10)[4] $-$ & 1.5819e+09(3.07e+09)[3] \\
 & 20 & 1.0583e+12(3.96e+11)[6] $-$ & 3.5753e+07(4.00e+07)[3] $+$ & 1.6355e+11(1.71e+11)[5] $-$ & 1.7005e+13(5.16e+12)[7] $-$ & 3.0900e+07(2.05e+07)[2] $+$ & \hl{9.0008e+04(1.98e+05)[1] $+$} & 1.1990e+10(1.49e+10)[4] \\
\hline
\multicolumn{2}{c}{Mean Rank} & 3.44 & 5.33 & 3.07 & 4.82 & 4.22 & 4.42 & 2.69 \\
\multicolumn{2}{c}{$+/-/\approx$} & 20/23/2 & 2/38/5 & 14/22/9 & 12/29/4 & 6/27/12 & 2/33/10  \\
\bottomrule
\end{tabular}
}
\label{tab:SOP-Results-all}
\end{table*}

We first compare R2SAEA-SOP with LSEA variants and EDA/LS on the LZG suite. Figure~\ref{fig:lzg_runtime_5d} shows the median convergence curves over 30 runs for $D=5$ (Additional numerical results are provided in the supplementary material). Overall, R2SAEA-SOP, powered by surrogate ReLLM-C1, achieves faster and more stable convergence than both the baseline EA and LSEA, indicating that a specialized relation LLM surrogate is more effective than prompting a general-purpose LLM for regression or classification within the same SAEA pipeline.

Table~\ref{tab:SOP-Results-all} reports the mean objective values on the combined LZG and YLL suites for $D=5,10,20$. R2SAEA-SOP attains the best performance on the majority of instances. In particular, R2SAEA-SOP outperforms FCPS and SAMSO on 27/45 and 33/45 cases, respectively. Compared to the relation-based DRSO, it achieves better results on 22/45 cases and comparable results on 9/45 cases. Against BO, R2SAEA-SOP performs better on 23/45 cases. In terms of mean rank, it ranks first among the seven algorithms, with a mean rank of 2.69. These results demonstrate that R2SAEA-SOP is competitive and robust across diverse landscapes under tight evaluation budgets.

\subsection{Empirical Studies in MOPs}
\label{subsec:experiments_mops}

\subsubsection{Algorithms for comparison}
To evaluate the proposed multi-objective framework (denoted as \textbf{R2SAEA-MOP}), we compare it with representative surrogate-assisted MOEAs spanning the three mainstream surrogate paradigms: classification, regression, and relation modeling. Specifically,
CPS-MOEA~\cite{zhang2015classification} and CSEA~\cite{pan2018classification} use KNN and neural networks, respectively, as classification surrogates to identify promising offspring.
K-RVEA~\cite{chugh2016surrogate} adopts a Gaussian process as regression surrogate to estimate objective values and couples it with reference vectors to maintain diversity.
We also include two relation-based approaches: REMO~\cite{hao2022expensive}, which learns pairwise preference relations with a neural network and applies a scoring strategy for selection, and PC-SAEA~\cite{tian2023pairwise}, which employs a probabilistic neural network with tailored selection mechanisms.
All baselines are run using the PlatEMO framework~\cite{tian2017platemo} with their default parameter settings to ensure a fair and reproducible comparison.

\subsubsection{Parameter Settings}
Following the expensive optimization setting, the maximum number of expensive evaluations is limited to $FEs=300$, and the population size is fixed at $N=30$. To test scalability across objective and decision spaces, we set the number of objectives to $M \in \{3,6,10\}$ and vary the decision dimension as $D\in\{5,10,30\}$. Each configuration is independently repeated 30 times.

We use the Inverted Generational Distance (IGD)~\cite{coello2005solving} as the primary performance indicator. Statistical significance is assessed via the Wilcoxon rank-sum test at the 0.05 level. In the result tables, the symbols ``$+$'', ``$-$'', and ``$\approx$'' indicate that a baseline performs significantly better than, significantly worse than, or statistically similar to R2SAEA-MOP, respectively. The best mean IGD value in each row is highlighted.

\subsubsection{Benchmark Problems}
We evaluate all algorithms on three widely used multi-objective benchmark suites: DTLZ~\cite{deb2005scalable} (DTLZ1--DTLZ7), WFG~\cite{huband2006review} (WFG1--WFG9), and MaF~\cite{cheng2017benchmark} (MaF1--MaF5). These benchmarks cover diverse Pareto-front geometries and search difficulties, providing a comprehensive testbed for expensive multi-objective optimization.

\definecolor{hl}{rgb}{0.75,0.75,0.75}
\sethlcolor{hl}

\begin{table*}[htbp]
\renewcommand{\arraystretch}{1.2}
\centering
\caption{Statistics of mean and standard deviation results obtained by five comparison algorithms on DTLZ test suites with M = 3, 6 10 and D = 5, 10, 30.}
\resizebox{\textwidth}{!}{%
\begin{tabular}{ccccccccc}
\toprule
Problem&$M$&$D$&CPS-MOEA&CSEA&K-RVEA&REMO&PC-SAEA&R2SAEA-MOP \\
\midrule
\multirow{8}{*}{DTLZ1}
    & 3 & 5 & 8.7388e+0(3.97e+0)[4] $-$ & 8.3268e+0(3.46e+0)[3] $-$ & 1.2952e+1(4.46e+0)[6] $-$ & 6.7390e+0(3.08e+0)[2] $-$ & 8.7929e+0(4.47e+0)[5] $-$ & \hl{3.6759e+0(2.37e+0)[1]} \\
    & 3 & 10 & 6.6358e+1(9.19e+0)[3] $-$ & 6.7623e+1(1.13e+1)[4] $-$ & 7.4204e+1(1.53e+1)[5] $-$ & 5.9098e+1(1.39e+1)[2] $-$ & 7.6135e+1(1.72e+1)[6] $-$ & \hl{3.2335e+1(1.28e+1)[1]} \\
    & 3 & 30 & 4.1273e+2(7.13e+1)[2] $-$ & 4.6932e+2(7.49e+1)[4] $-$ & 6.4035e+2(5.41e+1)[6] $-$ & 4.5629e+2(5.73e+1)[3] $-$ & 6.2992e+2(4.79e+1)[5] $-$ & \hl{3.5547e+2(6.38e+1)[1]} \\
    & 6 & 5 & 1.2761e-1(5.08e-3)[5] $\approx$ & 1.0617e-1(8.57e-3)[3] $+$ & \hl{7.6445e-2(2.51e-3)[1] $+$} & 1.0902e-1(1.23e-2)[4] $+$ & 8.2698e-2(5.52e-3)[2] $+$ & 1.3284e-1(1.12e-2)[6] \\
    & 6 & 10 & 4.2260e+1(1.34e+1)[6] $-$ & 2.2034e+1(6.33e+0)[3] $-$ & 2.7143e+1(7.49e+0)[4] $-$ & 1.7488e+1(5.86e+0)[2] $-$ & 3.3986e+1(1.06e+1)[5] $-$ & \hl{9.4570e+0(4.92e+0)[1]} \\
    & 6 & 30 & 4.7069e+2(5.37e+1)[6] $-$ & 3.5122e+2(5.55e+1)[3] $-$ & 4.6255e+2(4.66e+1)[4] $-$ & 2.9243e+2(6.09e+1)[2] $-$ & 4.6512e+2(4.13e+1)[5] $-$ & \hl{2.4589e+2(4.19e+1)[1]} \\
    & 10 & 10 & 1.1534e+0(1.22e+0)[6] $-$ & 3.1174e-1(9.98e-2)[2] $+$ & 3.5221e-1(1.07e-1)[3] $\approx$ & \hl{2.7077e-1(5.62e-2)[1] $+$} & 3.5674e-1(1.19e-1)[4] $\approx$ & 4.9988e-1(3.52e-1)[5] \\
    & 10 & 30 & 3.4929e+2(3.92e+1)[4] $-$ & 2.3798e+2(4.05e+1)[3] $-$ & 3.6658e+2(2.75e+1)[6] $-$ & 2.1983e+2(4.13e+1)[2] $-$ & 3.6544e+2(3.21e+1)[5] $-$ & \hl{1.9518e+2(3.75e+1)[1]} \\
\hline
\multirow{8}{*}{DTLZ2}
    & 3 & 5 & 1.4885e-1(9.69e-3)[5] $-$ & 1.5142e-1(1.04e-2)[6] $-$ & 9.8040e-2(5.75e-3)[3] $\approx$ & 1.0824e-1(1.36e-2)[4] $-$ & 9.7120e-2(8.46e-3)[2] $\approx$ & \hl{9.5381e-2(9.65e-3)[1]} \\
    & 3 & 10 & 2.8735e-1(3.42e-2)[5] $-$ & 3.1585e-1(2.70e-2)[6] $-$ & \hl{1.2869e-1(2.01e-2)[1] $+$} & 2.4327e-1(4.24e-2)[4] $-$ & 2.3763e-1(2.29e-2)[3] $-$ & 1.6506e-1(2.04e-2)[2] \\
    & 3 & 30 & 9.0069e-1(1.13e-1)[2] $-$ & 1.0852e+0(1.36e-1)[4] $-$ & 1.4672e+0(5.68e-2)[5] $-$ & 9.1556e-1(1.35e-1)[3] $-$ & 1.4778e+0(8.27e-2)[6] $-$ & \hl{6.5414e-1(1.32e-1)[1]} \\
    & 6 & 5 & 3.8981e-1(7.94e-3)[6] $-$ & 3.1645e-1(1.89e-2)[4] $\approx$ & 3.1369e-1(1.03e-2)[3] $\approx$ & 2.6917e-1(1.67e-2)[2] $+$ & \hl{2.6693e-1(1.28e-2)[1] $+$} & 3.1750e-1(2.93e-2)[5] \\
    & 10 & 10 & 6.8587e-1(3.45e-2)[6] $-$ & 6.3324e-1(3.27e-2)[4] $\approx$ & \hl{5.2681e-1(1.91e-2)[1] $+$} & 5.6857e-1(3.24e-2)[2] $+$ & 5.8319e-1(3.14e-2)[3] $+$ & 6.3767e-1(4.57e-2)[5] \\
    & 10 & 30 & 2.1788e+0(2.38e-1)[6] $-$ & 1.2482e+0(9.03e-2)[3] $-$ & 1.5639e+0(5.23e-2)[5] $-$ & 1.1404e+0(9.97e-2)[2] $-$ & 1.5424e+0(6.99e-2)[4] $-$ & \hl{1.0138e+0(9.68e-2)[1]} \\
    & 6 & 10 & 7.4033e-1(4.89e-2)[6] $-$ & 4.7057e-1(3.69e-2)[5] $-$ & \hl{3.6558e-1(1.60e-2)[1] $+$} & 3.9467e-1(3.90e-2)[2] $+$ & 4.3981e-1(2.43e-2)[4] $\approx$ & 4.2952e-1(4.54e-2)[3] \\
    & 6 & 30 & 2.3231e+0(2.71e-1)[6] $-$ & 1.2417e+0(1.25e-1)[3] $-$ & 1.5405e+0(6.68e-2)[4] $-$ & 1.0798e+0(1.48e-1)[2] $-$ & 1.5685e+0(6.25e-2)[5] $-$ & \hl{8.4993e-1(1.11e-1)[1]} \\
\hline
\multirow{8}{*}{DTLZ3}
    & 3 & 5 & 2.3965e+1(1.37e+1)[5] $-$ & 2.3278e+1(8.57e+0)[4] $-$ & 3.8411e+1(1.16e+1)[6] $-$ & 1.8624e+1(9.13e+0)[2] $-$ & 2.3092e+1(1.02e+1)[3] $-$ & \hl{9.1293e+0(4.51e+0)[1]} \\
    & 3 & 10 & 1.8592e+2(3.02e+1)[4] $-$ & 1.6788e+2(4.01e+1)[3] $-$ & 2.0617e+2(3.71e+1)[5] $-$ & 1.4060e+2(4.28e+1)[2] $-$ & 2.1447e+2(4.00e+1)[6] $-$ & \hl{9.2625e+1(2.75e+1)[1]} \\
    & 3 & 30 & 1.0809e+3(1.73e+2)[2] $\approx$ & 1.3561e+3(1.69e+2)[4] $-$ & 1.9567e+3(1.43e+2)[6] $-$ & 1.2632e+3(1.71e+2)[3] $-$ & 1.9179e+3(1.31e+2)[5] $-$ & \hl{1.0661e+3(1.59e+2)[1]} \\
    & 6 & 5 & 3.9349e-1(1.26e-2)[6] $-$ & 3.1893e-1(2.01e-2)[5] $\approx$ & 3.1235e-1(9.17e-3)[4] $\approx$ & 2.6809e-1(2.08e-2)[2] $+$ & \hl{2.6236e-1(1.04e-2)[1] $+$} & 3.1221e-1(3.03e-2)[3] \\
    & 10 & 10 & 9.9103e+0(8.44e+0)[6] $-$ & 1.1240e+0(3.32e-1)[2] $\approx$ & 1.2626e+0(3.31e-1)[4] $\approx$ & \hl{1.0342e+0(2.84e-1)[1] $+$} & 1.1667e+0(3.02e-1)[3] $\approx$ & 1.5532e+0(1.06e+0)[5] \\
    & 10 & 30 & 1.2765e+3(2.39e+2)[4] $-$ & 8.9401e+2(1.49e+2)[3] $-$ & 1.3316e+3(1.09e+2)[5] $-$ & 8.1988e+2(1.50e+2)[2] $-$ & 1.3740e+3(9.00e+1)[6] $-$ & \hl{6.8995e+2(1.47e+2)[1]} \\
    & 6 & 10 & 1.1979e+2(3.69e+1)[6] $-$ & 5.7720e+1(1.52e+1)[3] $-$ & 9.2775e+1(1.86e+1)[4] $-$ & 5.3167e+1(1.41e+1)[2] $-$ & 9.8893e+1(2.85e+1)[5] $-$ & \hl{3.0517e+1(1.14e+1)[1]} \\
    & 6 & 30 & 1.3953e+3(2.13e+2)[4] $-$ & 1.2220e+3(1.45e+2)[3] $-$ & 1.6545e+3(1.22e+2)[5] $-$ & 1.1140e+3(1.45e+2)[2] $-$ & 1.6822e+3(1.53e+2)[6] $-$ & \hl{9.0659e+2(1.70e+2)[1]} \\
\hline
\multirow{8}{*}{DTLZ4}
    & 3 & 5 & 3.5476e-1(6.35e-2)[4] $\approx$ & 2.3327e-1(4.37e-2)[3] $+$ & 1.7122e-1(5.19e-2)[2] $+$ & \hl{1.3387e-1(1.70e-2)[1] $+$} & 3.6904e-1(1.94e-1)[5] $\approx$ & 3.7862e-1(1.89e-1)[6] \\
    & 3 & 10 & 5.5758e-1(6.60e-2)[6] $\approx$ & 3.8628e-1(9.55e-2)[3] $+$ & 3.8338e-1(7.80e-2)[2] $+$ & \hl{2.1800e-1(2.61e-2)[1] $+$} & 4.7294e-1(1.71e-1)[4] $\approx$ & 4.9021e-1(1.30e-1)[5] \\
    & 3 & 30 & 1.2686e+0(1.42e-1)[4] $-$ & 9.8899e-1(1.65e-1)[3] $-$ & 1.8588e+0(8.22e-2)[6] $-$ & 8.7720e-1(9.63e-2)[2] $-$ & 1.8299e+0(9.57e-2)[5] $-$ & \hl{8.5064e-1(3.14e-1)[1]} \\
    & 6 & 5 & 4.9181e-1(1.48e-2)[5] $+$ & 4.1907e-1(3.93e-2)[3] $+$ & \hl{3.8973e-1(5.38e-2)[1] $+$} & 4.4829e-1(5.10e-2)[4] $+$ & 4.0440e-1(2.76e-2)[2] $+$ & 5.8026e-1(5.57e-2)[6] \\
    & 10 & 10 & 7.2374e-1(1.05e-2)[5] $+$ & 6.4839e-1(2.35e-2)[3] $+$ & 6.2908e-1(2.57e-2)[2] $+$ & 6.6126e-1(3.46e-2)[4] $+$ & \hl{6.1806e-1(1.42e-2)[1] $+$} & 8.0575e-1(4.75e-2)[6] \\
    & 10 & 30 & 2.0504e+0(1.95e-1)[6] $-$ & 1.2292e+0(1.05e-1)[3] $-$ & 1.7433e+0(7.39e-2)[4] $-$ & 1.1890e+0(8.95e-2)[2] $-$ & 1.7483e+0(5.06e-2)[5] $-$ & \hl{1.0675e+0(6.10e-2)[1]} \\
    & 6 & 10 & 7.1987e-1(3.96e-2)[6] $-$ & 5.3731e-1(4.00e-2)[3] $+$ & 5.2252e-1(5.59e-2)[2] $+$ & \hl{5.1758e-1(4.97e-2)[1] $+$} & 6.0874e-1(2.66e-2)[4] $+$ & 6.6006e-1(7.75e-2)[5] \\
    & 6 & 30 & 1.8910e+0(2.41e-1)[6] $-$ & 1.2163e+0(9.94e-2)[3] $-$ & 1.8441e+0(7.98e-2)[4] $-$ & 1.0974e+0(1.44e-1)[2] $-$ & 1.8721e+0(8.02e-2)[5] $-$ & \hl{9.6200e-1(1.01e-1)[1]} \\
\hline
\multirow{8}{*}{DTLZ5}
    & 3 & 5 & 3.1759e-2(5.79e-3)[4] $-$ & 5.9459e-2(1.32e-2)[6] $-$ & \hl{2.0367e-2(4.92e-3)[1] $\approx$} & 3.0393e-2(8.21e-3)[3] $-$ & 3.3595e-2(6.86e-3)[5] $-$ & 2.0561e-2(4.59e-3)[2] \\
    & 3 & 10 & 1.5670e-1(3.90e-2)[4] $-$ & 2.2046e-1(3.55e-2)[6] $-$ & \hl{5.7034e-2(1.51e-2)[1] $\approx$} & 1.4315e-1(3.85e-2)[3] $-$ & 1.6506e-1(2.93e-2)[5] $-$ & 6.0403e-2(1.18e-2)[2] \\
    & 3 & 30 & 7.5108e-1(1.72e-1)[2] $-$ & 9.7829e-1(1.27e-1)[4] $-$ & 1.3941e+0(1.04e-1)[5] $-$ & 8.7853e-1(1.42e-1)[3] $-$ & 1.4193e+0(1.04e-1)[6] $-$ & \hl{5.4811e-1(1.15e-1)[1]} \\
    & 6 & 5 & 2.2646e-2(3.76e-3)[6] $-$ & 4.0936e-3(3.31e-4)[3] $+$ & 7.3209e-3(6.41e-4)[4] $\approx$ & 3.9566e-3(3.88e-4)[2] $+$ & \hl{3.6414e-3(2.12e-4)[1] $+$} & 7.5694e-3(4.78e-4)[5] \\
    & 10 & 10 & 8.4882e-2(1.81e-2)[6] $-$ & 1.4321e-2(1.90e-3)[2] $+$ & \hl{1.2504e-2(1.91e-3)[1] $+$} & 1.4896e-2(2.88e-3)[3] $+$ & 1.5393e-2(1.89e-3)[4] $+$ & 2.4518e-2(5.03e-3)[5] \\
    & 10 & 30 & 1.7684e+0(4.45e-1)[6] $-$ & 7.0432e-1(1.18e-1)[3] $-$ & 9.8892e-1(6.60e-2)[4] $-$ & 6.5286e-1(1.09e-1)[2] $-$ & 1.0144e+0(6.49e-2)[5] $-$ & \hl{4.2631e-1(8.71e-2)[1]} \\
    & 6 & 10 & 2.3812e-1(6.05e-2)[6] $-$ & 1.1074e-1(2.04e-2)[4] $-$ & \hl{3.3314e-2(7.87e-3)[1] $+$} & 7.5326e-2(2.26e-2)[3] $-$ & 1.2560e-1(1.59e-2)[5] $-$ & 4.5555e-2(1.53e-2)[2] \\
    & 6 & 30 & 2.0709e+0(4.22e-1)[6] $-$ & 9.1795e-1(1.36e-1)[3] $-$ & 1.2453e+0(7.89e-2)[4] $-$ & 8.4786e-1(1.34e-1)[2] $-$ & 1.2489e+0(8.16e-2)[5] $-$ & \hl{4.7600e-1(9.96e-2)[1]} \\
\hline
\multirow{8}{*}{DTLZ6}
    & 3 & 5 & 7.3064e-1(3.41e-1)[3] $\approx$ & 8.4457e-1(2.86e-1)[5] $\approx$ & \hl{2.8262e-1(1.56e-1)[1] $+$} & 3.5428e-1(2.55e-1)[2] $+$ & 1.4728e+0(3.68e-1)[6] $-$ & 7.7019e-1(3.97e-1)[4] \\
    & 3 & 10 & 3.0834e+0(5.86e-1)[2] $+$ & 4.7934e+0(4.95e-1)[5] $\approx$ & \hl{2.3602e+0(4.54e-1)[1] $+$} & 4.1246e+0(6.65e-1)[3] $\approx$ & 5.8658e+0(4.11e-1)[6] $-$ & 4.3819e+0(9.18e-1)[4] \\
    & 3 & 30 & \hl{1.5248e+1(1.78e+0)[1] $+$} & 2.1487e+1(9.15e-1)[3] $\approx$ & 2.4155e+1(1.96e-1)[5] $-$ & 2.1515e+1(8.71e-1)[4] $-$ & 2.4191e+1(1.64e-1)[6] $-$ & 2.0940e+1(1.16e+0)[2] \\
    & 6 & 5 & 2.2162e-2(3.01e-3)[6] $-$ & 4.2231e-3(3.55e-4)[3] $+$ & 7.4562e-3(6.16e-4)[5] $\approx$ & 4.0048e-3(3.13e-4)[2] $+$ & \hl{3.5679e-3(2.67e-4)[1] $+$} & 7.3567e-3(4.72e-4)[4] \\
    & 10 & 10 & 2.6458e-1(1.22e-1)[5] $-$ & 1.0700e-1(1.17e-1)[3] $\approx$ & \hl{6.5720e-2(2.76e-2)[1] $\approx$} & 8.0111e-2(6.02e-2)[2] $\approx$ & 3.9721e-1(1.75e-1)[6] $-$ & 1.9070e-1(2.37e-1)[4] \\
    & 10 & 30 & \hl{1.3958e+1(1.65e+0)[1] $+$} & 1.6095e+1(8.53e-1)[3] $\approx$ & 1.8011e+1(1.95e-1)[5] $-$ & 1.6522e+1(7.53e-1)[4] $-$ & 1.8053e+1(1.80e-1)[6] $-$ & 1.6073e+1(9.52e-1)[2] \\
    & 6 & 10 & 2.6096e+0(7.57e-1)[5] $\approx$ & 2.6027e+0(4.04e-1)[4] $\approx$ & \hl{1.3160e+0(4.15e-1)[1] $+$} & 2.4378e+0(6.59e-1)[2] $\approx$ & 3.6585e+0(3.15e-1)[6] $-$ & 2.5296e+0(8.63e-1)[3] \\
    & 6 & 30 & \hl{1.7070e+1(1.58e+0)[1] $+$} & 1.9545e+1(9.44e-1)[2] $\approx$ & 2.1529e+1(2.10e-1)[5] $-$ & 1.9798e+1(7.75e-1)[4] $\approx$ & 2.1569e+1(2.04e-1)[6] $-$ & 1.9580e+1(8.93e-1)[3] \\
\hline
\multirow{7}{*}{DTLZ7}
    & 3 & 5 & 7.0939e-1(3.19e-1)[4] $+$ & 4.3861e-1(1.54e-1)[3] $+$ & \hl{1.1389e-1(9.58e-3)[1] $+$} & 1.9136e-1(8.93e-2)[2] $+$ & 8.3449e-1(3.93e-1)[5] $\approx$ & 8.3909e-1(2.14e-1)[6] \\
    & 3 & 10 & 3.8825e+0(1.39e+0)[6] $-$ & 1.5988e+0(7.31e-1)[3] $\approx$ & \hl{1.3194e-1(1.26e-2)[1] $+$} & 4.8750e-1(2.83e-1)[2] $+$ & 3.1286e+0(1.13e+0)[5] $-$ & 1.7450e+0(4.71e-1)[4] \\
    & 3 & 30 & 7.9752e+0(8.20e-1)[4] $-$ & 5.0050e+0(1.00e+0)[3] $\approx$ & 8.9076e+0(4.35e-1)[6] $-$ & \hl{3.7188e+0(9.40e-1)[1] $+$} & 8.8027e+0(5.02e-1)[5] $-$ & 4.8177e+0(1.03e+0)[2] \\
    & 6 & 10 & 3.9075e+0(2.50e+0)[4] $+$ & 3.6323e+0(1.20e+0)[3] $+$ & \hl{6.3228e-1(5.41e-2)[1] $+$} & 2.4940e+0(6.18e-1)[2] $+$ & 7.1708e+0(2.16e+0)[6] $-$ & 4.7195e+0(1.13e+0)[5] \\
    & 6 & 30 & 1.4403e+1(2.49e+0)[2] $\approx$ & 1.4572e+1(2.05e+0)[3] $\approx$ & 1.7922e+1(1.02e+0)[5] $-$ & \hl{1.1952e+1(1.47e+0)[1] $+$} & 1.7951e+1(1.39e+0)[6] $-$ & 1.5048e+1(2.15e+0)[4] \\
    & 10 & 10 & 1.9393e+0(2.85e-1)[4] $+$ & 2.1270e+0(3.77e-1)[5] $+$ & \hl{1.0259e+0(3.39e-2)[1] $+$} & 1.9329e+0(3.11e-1)[3] $+$ & 1.7461e+0(3.75e-1)[2] $+$ & 3.1705e+0(7.55e-1)[6] \\
    & 10 & 30 & 2.3389e+1(3.57e+0)[2] $+$ & 2.6745e+1(3.00e+0)[4] $\approx$ & 2.9383e+1(2.31e+0)[6] $-$ & \hl{2.1888e+1(3.28e+0)[1] $+$} & 2.9303e+1(2.30e+0)[5] $-$ & 2.6157e+1(3.24e+0)[3] \\
\hline
\multicolumn{3}{c}{Mean Rank}&4.49&3.53&3.38&2.33&4.44&2.84\\
\multicolumn{3}{c}{$+/-/\approx$}&10/38/7&13/27/15&18/27/10&23/28/4&11/37/7&\\
\bottomrule
\end{tabular}
}
\label{tab:DTLZ_Summary}
\end{table*}


\subsubsection{Empirical Results}
The statistical results of the mean IGD values attained by five algorithms on DTLZ1-DTLZ7 with D = 5, 10, 30 and M = 3, 6, 10 are presented in Table~\ref{tab:DTLZ_Summary}. As evidenced by the results, R2SAEA-MOP outperforms the other algorithms in most test instances. More specifically, compared to the classification-based CPS-MOEA and CSEA, R2SAEA-MOP achieved superior results in 38 and 27 out of 55 test cases, respectively. When compared to the regression-based K-RVEA, R2SAEA-MOP achieved superior results in 27 out of 55 test cases, respectively. And, compared to relation-based PC-SAEA and REMO, R2SAEA-MOP yielded superior results in 37 and 28 out of 55 test cases. In terms of average rank, R2SAEA-MOP attains 2.84, ranking first. These findings suggest that R2SAEA-MOP outperforms the peer multi-objective SAEAs on the DTLZ test suite. Due to space constraints, detailed experimental results for WFG and MaF are provided in the supplementary materials.

Notably, the ReLLM-C2 model used in R2SAEA-MOP is trained solely on a few functions in SOP, yet it exhibits effective reasoning capability on MOP problems, indicating the generalizability of relational reasoning and the effectiveness of the RL stage.

\subsection{Comparison with General LLMs}
\label{subsec:experiments_general_models}

This subsection evaluates the relation prediction capability of general-purpose LLMs and our models (ReLLM-C1/ReLLM-C2). Directly assessing closed-source LLMs within full evolutionary runs is impractical due to API cost and latency, and the stochasticity of EA trajectories further complicates repeated controlled comparisons. 
Therefore, we adopt an offline evaluation protocol based on static populations sampled from evolutionary trajectories.

We construct the offline benchmark from the four classical LZG functions. For each function, we run a genetic algorithm (GA)~\cite{holland1992adaptation} with population size $N=100$ and decision dimension $D=5$ for 100 generations. To emulate the population distributions encountered at different phases of optimization, we sample three representative stages (generations 10, 50, and 90). At each stage, we draw 30 individuals from the parent population as the context set (i.e., $\mathcal{D}_{\mathrm{train}}$ in our prompt construction) and 30 individuals from the offspring population as the query candidates (i.e., $\mathcal{D}_{\mathrm{test}}$). This setting matches the anchor-based iterative inference pipeline used in the EA (Section~\ref{subsec:AnchorBasedEngineering}) while enabling controlled, repeatable evaluation.

We consider leading closed-source models, including GPT-4o~\cite{hurst2024gpt}, Gemini-2.5-Pro~\cite{comanici2025gemini}, and Claude-4 Sonnet~\cite{anthropic2025claude4}, as well as strong open-source models, including Qwen3-235B~\cite{yang2025qwen3} and Qwen2.5-7B~\cite{qwen2025qwen25technicalreport}. Qwen2.5-7B is also used as our base model, enabling a direct comparison between pre- and post- RL fine-tuning.

\begin{figure}[ht!]
    \centering
    \includegraphics[width=1\linewidth]{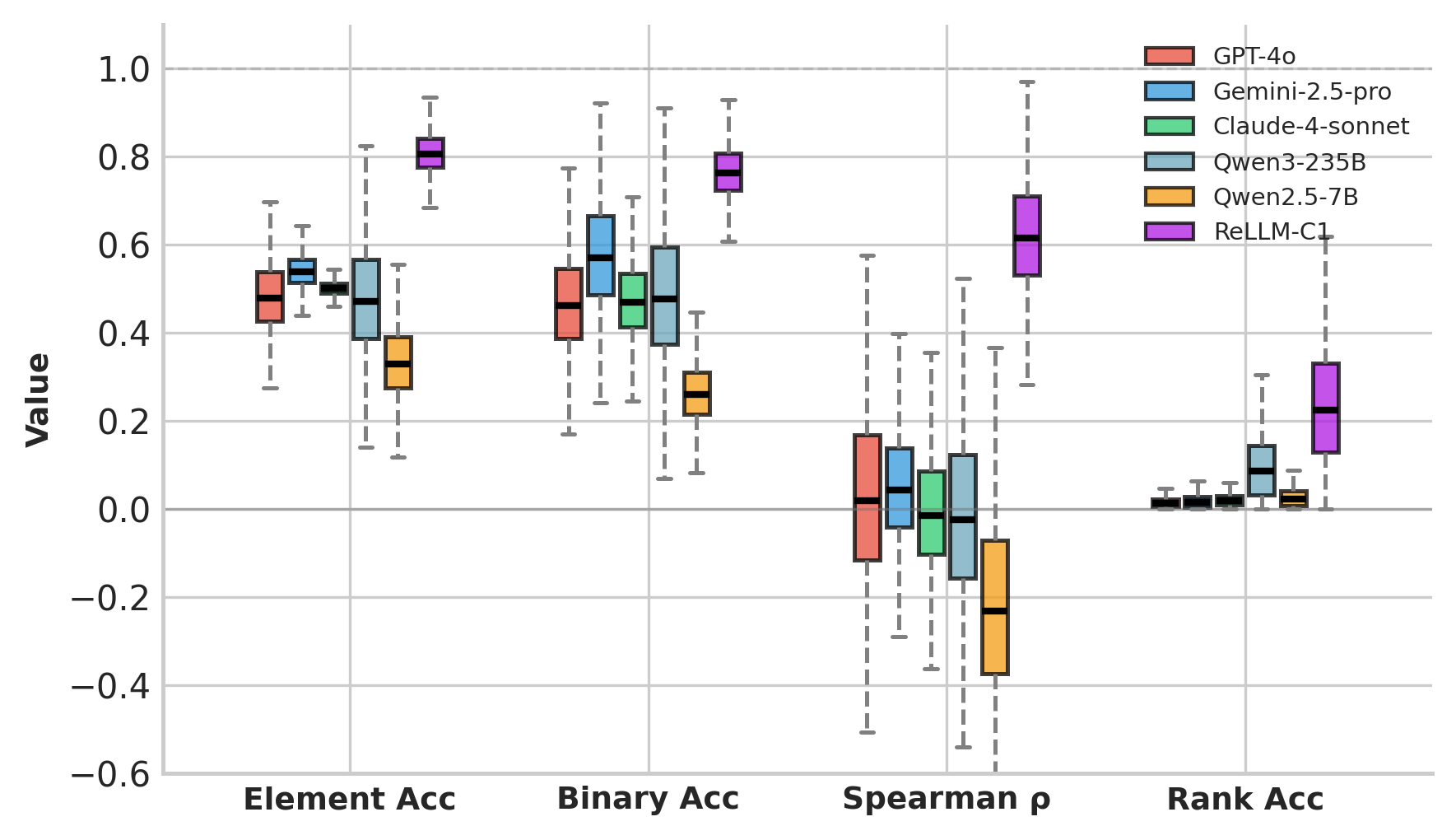}
    \caption{Offline evaluation of ReLLM-C1 and general LLMs under the C1 criterion.}
    \label{fig:offline-C1}
\end{figure}

\begin{figure}[ht!]
    \centering
    \includegraphics[width=1\linewidth]{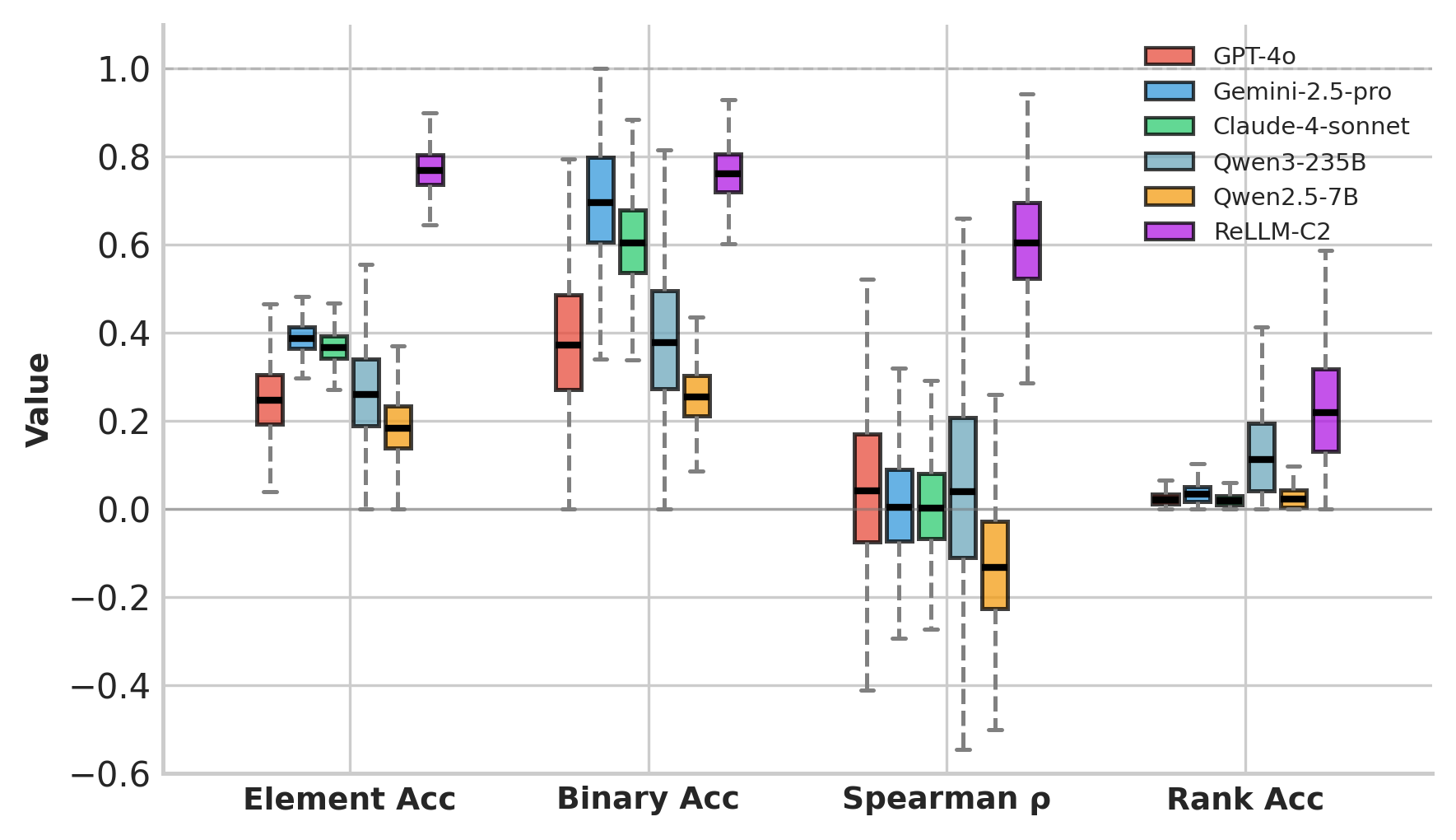}
    \caption{Offline evaluation of ReLLM-C2 and general LLMs under the C2 criterion.}
    \label{fig:offline-C2}
\end{figure}

We report four metrics to reflect both local relation correctness and downstream selection usefulness: 
\textbf{Element Acc} (pairwise label accuracy), 
\textbf{Binary Acc} (accuracy of selecting the better half of candidates, mimicking model-assisted offspring screening), and two rank-consistency metrics between the vote-aggregated ordering and the ground-truth ordering: \textbf{Rank Acc} and \textbf{Spearman's $\rho$}. Formal definitions are provided in the supplementary material.

As shown in Fig.~\ref{fig:offline-C1}--Fig.~\ref{fig:offline-C2}, ReLLM-C1 and ReLLM-C2 consistently outperform their pre-fine-tuning base model across all metrics, demonstrating that GRPO effectively specializes the LLM for relation inference. 
Moreover, the proposed models achieve a clear advantage over both closed-source and open-source general LLMs on this task. 
In particular, under Spearman's $\rho$, most general LLMs yield correlations near zero (or even negative), indicating that without task-specific training they cannot reliably infer the pairwise relations required to produce a meaningful ranking after voting-based aggregation. These results explain the performance gains observed when integrating ReLLM into SAEAs in Sections~\ref{subsec:experiments_sops} and~\ref{subsec:experiments_mops}.

\subsection{Data Scalability Experiment}
\label{subsec:experiments_data_scaled}

\begin{figure}[ht!]
    \includegraphics[width=1\linewidth]{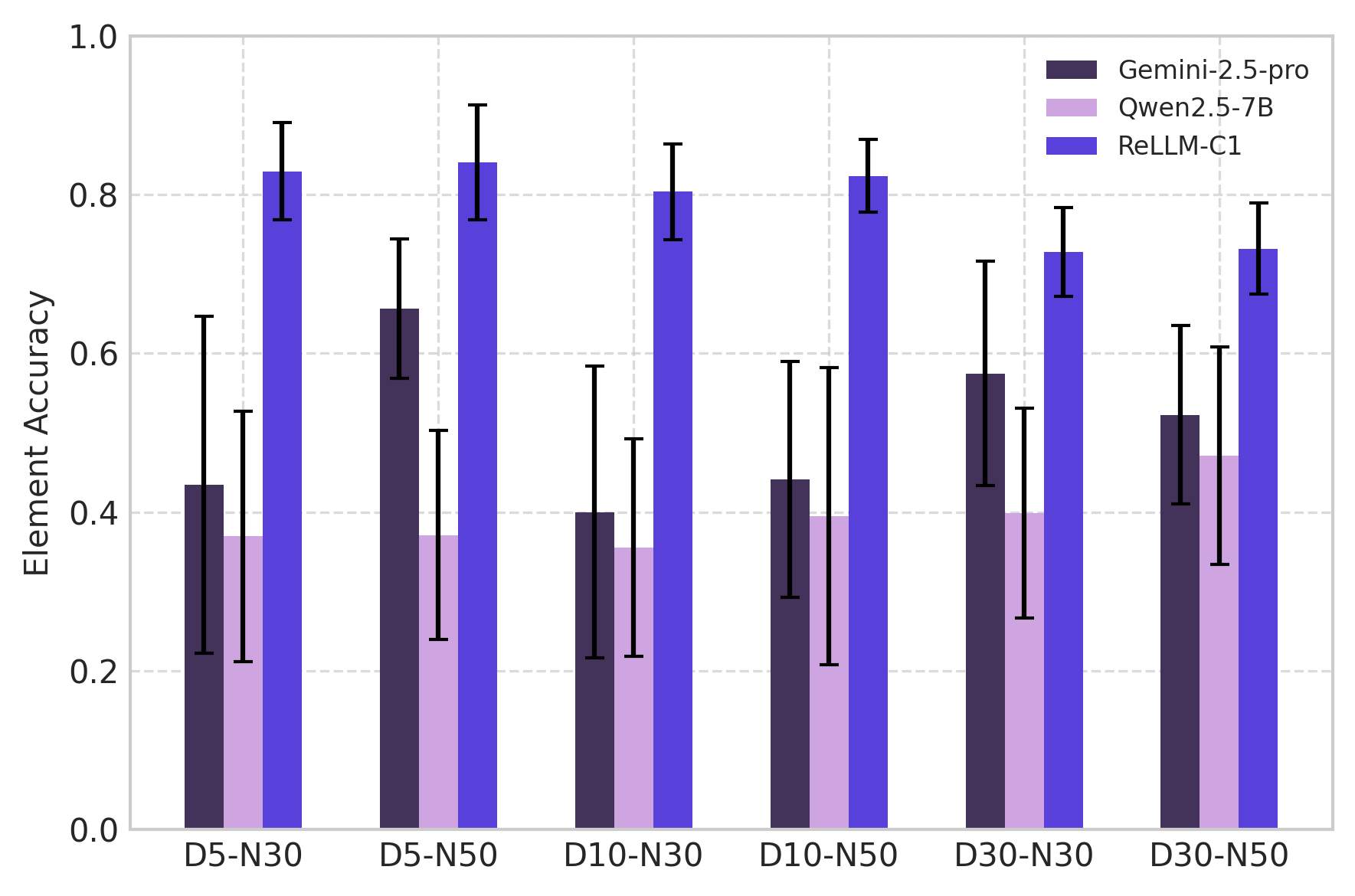}
    \caption{Element Acc ($\mathrm{acc}_\mathrm{ele}$) on the offline test set under different decision dimensions ($D$) and subsample sizes ($N$).}
    \label{fig:scaled_elem_acc}
\end{figure}

\begin{figure}[ht!]
    \includegraphics[width=1\linewidth]{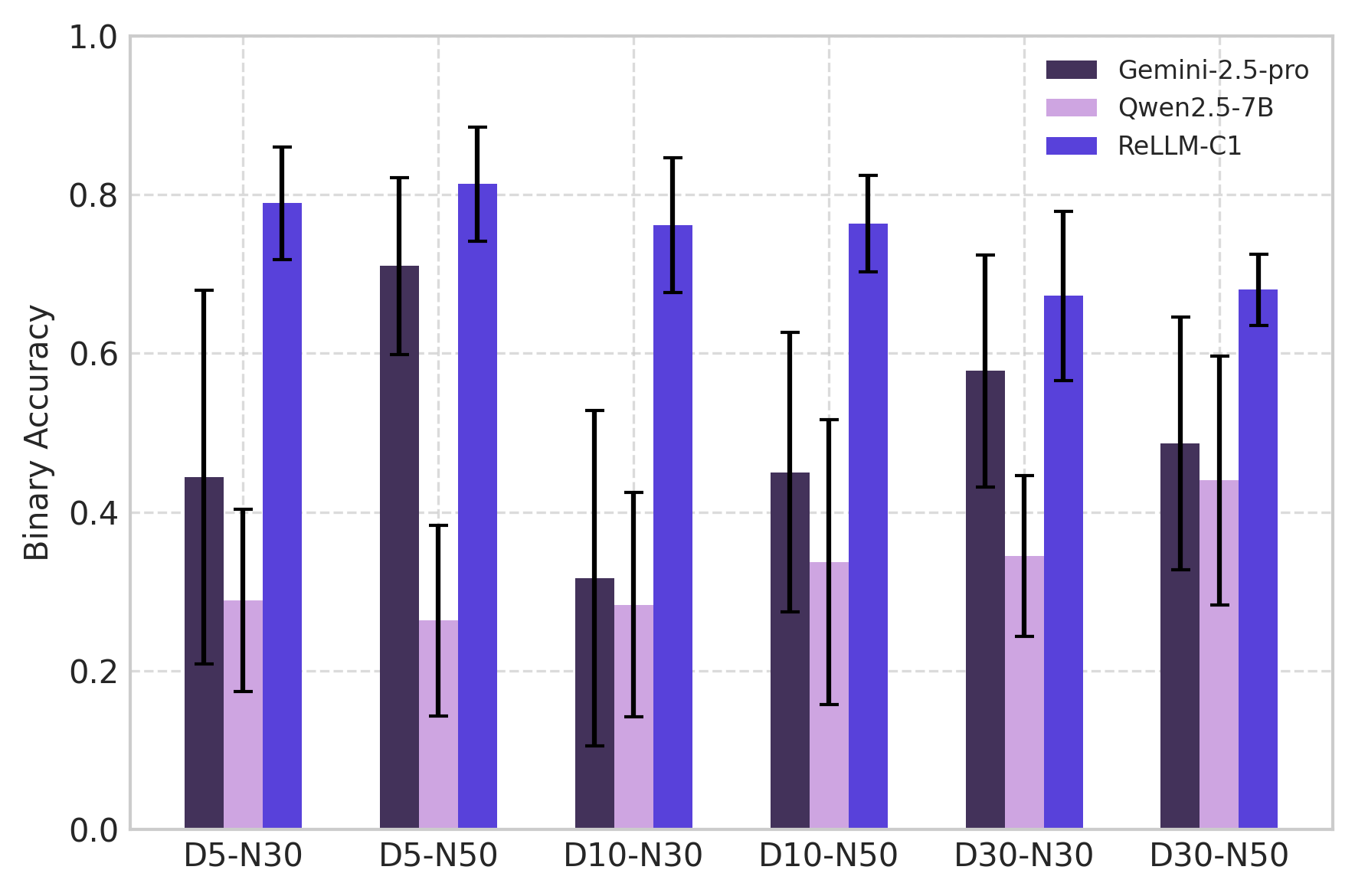}
    \caption{Binary Acc ($\mathrm{acc}_\mathrm{bin}$) on the offline test set under different decision dimensions ($D$) and subsample sizes ($n$).}
    \label{fig:scaled_bin_acc}
\end{figure}

This subsection examines how well the proposed relational surrogate generalizes when the data scale changes, including both the decision-space dimension~($D$) and the number of context/query samples~($N$). We focus on ReLLM-C1, which is RL fine-tuned using trajectories with subsample size $n=30$ and dimension $D=5$ (Section~\ref{subsec:proposed_method_RL}). We then test the model on the offline protocol (Section~\ref{subsec:experiments_general_models}) under two subsample sizes, $N\in\{30,50\}$, and three decision dimensions, $D\in\{5,10,30\}$. 
As baselines, we include the pre-fine-tuning base model Qwen2.5-7B and Gemini-2.5-Pro, which performs best among the general LLMs in Section~\ref{subsec:experiments_general_models}. We report \textbf{Element Acc} ($\mathrm{acc}_\mathrm{ele}$), measuring pairwise relation correctness, and \textbf{Binary Acc} ($\mathrm{acc}_\mathrm{bin}$), measuring the accuracy of selecting the better half of candidates (i.e., a proxy for model-assisted offspring screening).
As shown in Fig.~\ref{fig:scaled_elem_acc}--Fig.~\ref{fig:scaled_bin_acc}, increasing the decision dimension $D$ leads to a mild degradation in accuracy, while increasing the subsample size $N$ consistently improves both metrics, suggesting that richer context helps the relational inference. Importantly, ReLLM-C1 remains the top-performing model across all tested settings, consistently outperforming both Qwen2.5-7B and Gemini-2.5-Pro. This indicates that the GRPO-specialized relational surrogate retains strong robustness under data scale shifts beyond its training condition.


\subsection{The Impact of Normalization Strategies}
\label{subsec:experiments_normalization}

Normalization is required in our RL fine-tuning and inference pipeline, since decision variables originate from different benchmark functions and different evolutionary stages, and therefore may have heterogeneous value ranges. We denote the context set as $\mathcal{D}_{\mathrm{train}}$ and the candidate set as $\mathcal{D}_{\mathrm{test}}$. We study two normalization strategies that differ in scope:

\begin{enumerate}
    \item Global normalization (dataset-level):
    We compute the min--max statistics once using the entire RL training corpus collected from multiple functions and multiple generations, and then apply the same normalization to all training instances. Formally, let $\mathcal{D}_{\mathrm{RL}}$ denote the union of all decision vectors appearing in the RL dataset (including both context vectors from $\mathcal{D}_{\mathrm{train}}$ and candidates from $\mathcal{D}_{\mathrm{test}}$ across all collected trajectories). For any decision vector $\mathbf{z}$, we apply
    \begin{equation}
        \mathrm{Norm}_{\mathrm{global}}(\mathbf{z})
        =
        \frac{\mathbf{z}-\min(\mathcal{D}_{\mathrm{RL}})}
        {\max(\mathcal{D}_{\mathrm{RL}})-\min(\mathcal{D}_{\mathrm{RL}})},
    \end{equation}
    where $\min(\mathcal{D}_{\mathrm{RL}})$ and $\max(\mathcal{D}_{\mathrm{RL}})$ are computed dimension-wise. 
    This strategy enforces that all RL training data are represented under a shared normalization space.
    \item Local normalization (prompt-level):
    We compute min--max statistics separately for each inference instance (i.e., each prompt). Concretely, for a given prompt constructed from the current context $\mathcal{D}_{\mathrm{train}}$ and the current candidate set $\mathcal{D}_{\mathrm{test}}$, we define $\mathcal{D}=\mathcal{D}_{\mathrm{train}}\cup\mathcal{D}_{\mathrm{test}}$ and normalize each vector $\mathbf{z}\in\mathcal{D}$ via
    \begin{equation}
        \mathrm{Norm}_{\mathrm{local}}(\mathbf{z})
        =
        \frac{\mathbf{z}-\min(\mathcal{D})}
        {\max(\mathcal{D})-\min(\mathcal{D})},
    \end{equation}
    where $\min(\mathcal{D})$ and $\max(\mathcal{D})$ are computed dimension-wise using only the data that appear in this prompt.
\end{enumerate}

Global normalization is common in standard ML pipelines because it provides a consistent feature scale across the entire dataset. However, our task is relational and is solved via in-context inference: the model primarily needs to compare candidates against the current context in each prompt. Local (prompt-level) normalization can therefore be advantageous because it (i) amplifies the relative differences within the current $\mathcal{D}_{\mathrm{train}}$--$\mathcal{D}_{\mathrm{test}}$ pair, and (ii) reduces unintended dependence on absolute ranges that may vary across functions or evolutionary stages, encouraging the model to rely on contextual comparisons rather than dataset-specific scale cues.
We fine-tune and evaluate models under both normalization schemes. As shown in Table~\ref{tab:normalization_comparison}, local (prompt-level) normalization consistently yields better generalization and higher prediction accuracy on multiple test sets, supporting its use in our relational LLM surrogate.

\begin{table}[ht!]
\centering
\small
\setlength{\tabcolsep}{2pt}
\caption{Performance of Local normalization and Global normalization processing strategies.}
\label{tab:normalization_comparison}
\begin{tabular}{@{}lcccc@{}}
\toprule
& \multicolumn{2}{c}{\textbf{C1}} & \multicolumn{2}{c}{\textbf{C2}} \\
\cmidrule(lr){2-3} \cmidrule(lr){4-5}
\multirow{-2}{*}{\textbf{Metric}} & \textbf{Local} & \textbf{Global} & \textbf{Local} & \textbf{Global} \\
\midrule
\textbf{Element Acc}     & \textbf{0.815±0.065} & 0.724±0.081 & \textbf{0.777±0.066} & 0.734±0.079 \\
\textbf{Binary Acc}      & \textbf{0.775±0.084} & 0.725±0.085 & \textbf{0.772±0.085} & 0.725±0.120 \\
\textbf{Spearman $\rho$} & \textbf{0.639±0.179} & 0.441±0.248 & \textbf{0.627±0.171} & 0.349±0.258 \\
\textbf{Rank Acc}        & \textbf{0.250±0.200} & 0.064±0.063 & \textbf{0.244±0.186} & 0.222±0.233 \\
\bottomrule
\end{tabular}
\end{table}

\subsection{Quantification and deployment of performance}
\label{subsec:experiments_quantification}

A key practical advantage of using LLMs as surrogates is that, once specialized, they can operate in an inference-only manner inside the SAEA loop, avoiding per-generation retraining and potentially enabling hardware-friendly deployment. This subsection therefore studies how model scaling and quantization affect both relational prediction accuracy and inference speed.

Following the RL pipeline in Section~\ref{subsec:proposed_method_RL}, we fine-tune a smaller base model, Qwen2.5-3B-Instruct, to obtain a compact relational surrogate. We then apply 4-bit and 8-bit quantization (denoted as Q4 and Q8) using the \emph{K-quants} algorithm~\cite{llamacpp}, producing lightweight variants suitable for deployment. All quantized models are evaluated on the same offline test set described in Section~\ref{subsec:experiments_general_models}. Table~\ref{tab:quantification_performance} summarizes the results relative to the full-precision 3B and 7B instruct baselines across multiple accuracy metrics.
Overall, the RL specialization remains effective after downsizing: the fine-tuned 3B model achieves performance close to the 7B variant and substantially improves upon its pre-fine-tuning counterpart. Moreover, Q4 quantization introduces only marginal accuracy degradation, while still outperforming the full-precision base (non-fine-tuned) models, indicating that the learned relational capability is largely preserved under low-bit quantization.

\begin{table*}[ht!]
\centering
\small
\setlength{\tabcolsep}{5pt}
\caption{Performance of C1-Trained Models across Varying Parameter Sizes and Quantization Precisions on the Offline Test Set.}
\label{tab:quantification_performance}
\begin{tabular}{@{}lcccccc@{}}
\toprule
& \multicolumn{3}{c}{\textbf{3B}} & \multicolumn{3}{c}{\textbf{7B}} \\
\cmidrule(lr){2-4} \cmidrule(lr){5-7}
\multirow{-2}{*}{\textbf{Metric}} & \textbf{Qwen2.5-instruct} & \textbf{ReLLM-C1} & \textbf{ReLLM-C1-Q4KM} & \textbf{Qwen2.5-instruct} & \textbf{ReLLM-C1} & \textbf{ReLLM-C1-Q4KM} \\
\midrule
\textbf{Element Acc}     & 0.528±0.054[3] & \textbf{0.821±0.061[1]} & 0.778±0.080[2] & 0.345±0.114[3] & \textbf{0.815±0.065[1]} & 0.815±0.068[2] \\
\textbf{Binary Acc}      & 0.450±0.127[3] & \textbf{0.803±0.090[1]} & 0.750±0.129[2] & 0.272±0.095[3] & \textbf{0.775±0.084[1]} & 0.769±0.079[2] \\
\textbf{Spearman $\rho$} & -0.144±0.276[3] & \textbf{0.600±0.171[1]} & 0.558±0.184[2] & -0.192±0.302[3] & \textbf{0.639±0.179[1]} & 0.614±0.186[2] \\
\textbf{Rank Acc}        & 0.033±0.036[3] & \textbf{0.217±0.140[1]} & 0.156±0.144[2] & 0.028±0.033[3] & \textbf{0.250±0.200[1]} & 0.139±0.133[2] \\
\midrule
\textbf{Mean Rank}       & 3.0 & \textbf{1.0} & 2.0 & 3.0 & \textbf{1.0} & 2.0 \\
\bottomrule
\end{tabular}
\end{table*}

To assess practical deployment, we benchmark inference throughput on four platforms: NVIDIA H100, RTX~4090, Apple M3~Max, and an embedded edge device (Jetson Orin NX). Since our relation inference is typically \emph{long-input, short-output} (anchor-based prompts with compact JSON labels), we report two complementary metrics: Total-Tokens/s (TTs), measuring end-to-end throughput for input plus output tokens, and Completion-Tokens/s (CTs), measuring the generation speed of output tokens only. These metrics better reflect user-perceived latency and deployment cost.

\begin{figure}[ht!]
    \includegraphics[width=1.0\linewidth]{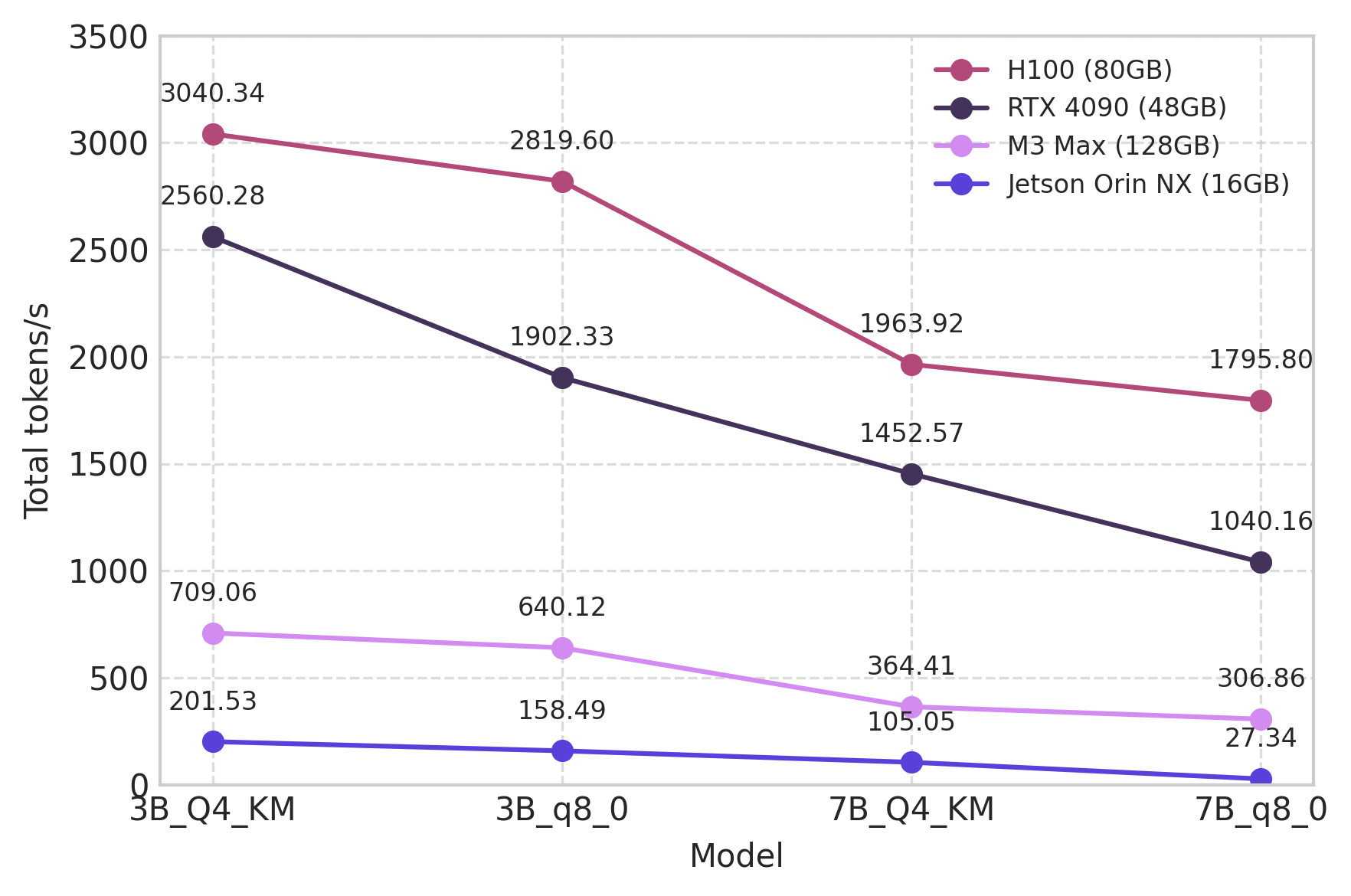}
    \caption{Total tokens per second~(TTs) by Model Size and Hardware}
    \label{fig:total_tokens}
\end{figure}

\begin{figure}[ht!]
    \includegraphics[width=1.0\linewidth]{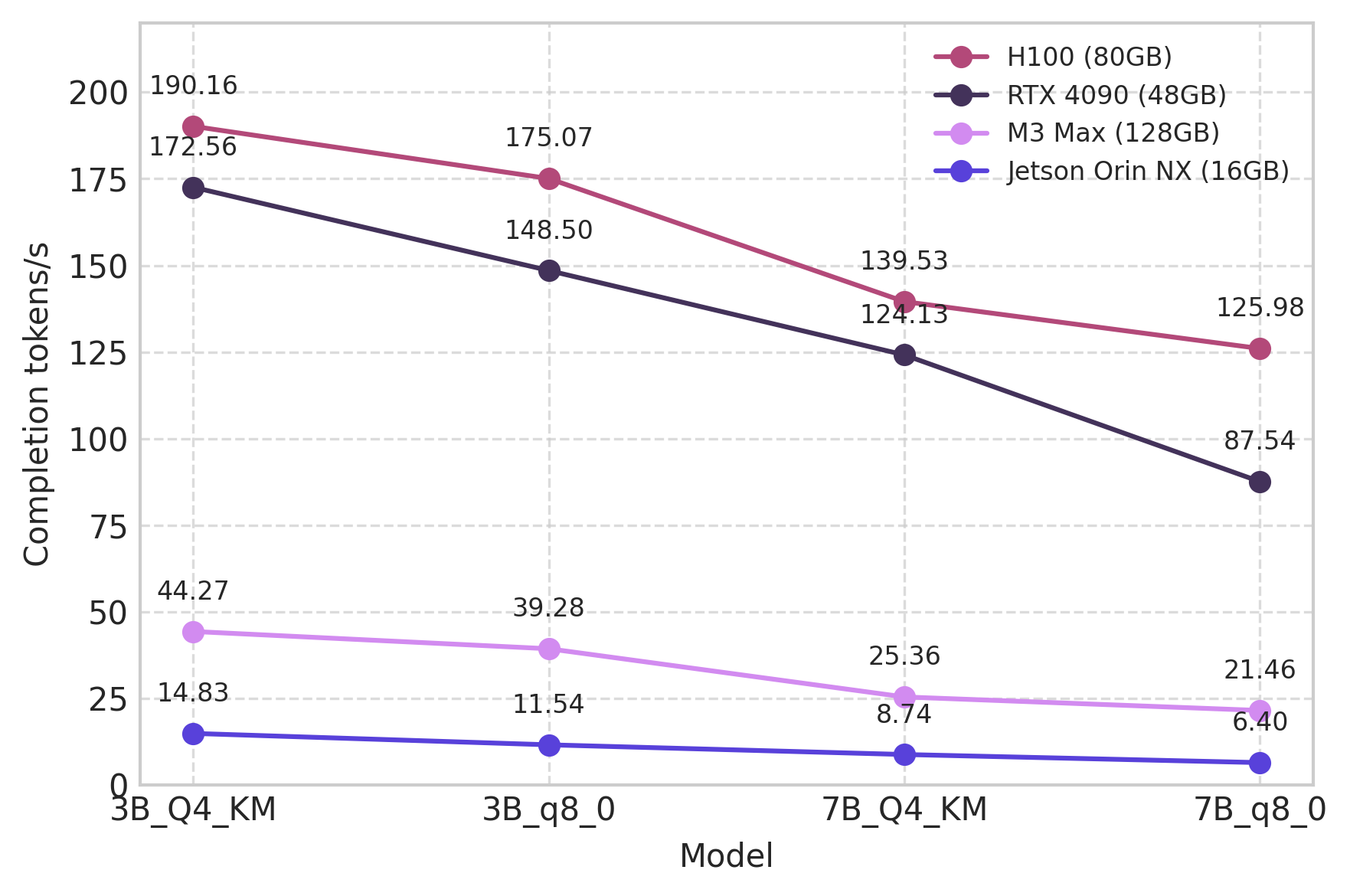}
    \caption{Completion tokens per second~(CTs) by Model Size and Hardware}
    \label{fig:comp_tokens}
\end{figure}

As shown in Fig.~\ref{fig:total_tokens} and Fig.~\ref{fig:comp_tokens}, the Jetson Orin NX (16GB) provides usable throughput despite strict power and memory constraints. In particular, with the 3B-Q4-KM model it achieves 201.53 TTs and 14.83 CTs. While slower than data-center GPUs, this performance suggests that, with appropriate model scaling and quantization, edge devices can support near-real-time relational inference, offering a practical pathway for deploying LLM-based surrogates in resource-limited scenarios.

\section{Conclusion}
\label{sec:conclusion}

This paper proposes a reinforcement-learning-driven framework for incorporating LLMs into expensive optimization as relational surrogates. We present a unified three-stage pipeline of relational data construction, anchor-based context engineering via iterative inference, and voting-based scoring, which transforms pairwise relation predictions into actionable scalar scores for model-assisted selection in SAEAs. To make this paradigm practical, we further fine-tune compact LLMs with GRPO. GRPO removes the need for an explicit critic and provides stable learning signals through group-wise reward normalization, improving relational reasoning while reducing reliance on costly frontier LLMs. Extensive experiments on both single-objective and multi-objective benchmarks show that the proposed method consistently outperforms strong SOTA baselines and general-purpose LLMs. The key contribution is enabling reliable zero-shot relation inference inside the evolutionary loop, thereby eliminating the iterative surrogate retraining required by conventional SAEAs and offering a feasible route toward hardware-friendly, inference-only surrogate deployment. We also release a suite of relation-reasoning LLM surrogates, which, to the best of our knowledge, are the first models explicitly fine-tuned for surrogate-assisted evolutionary optimization.

Future work will investigate context compression under strict token budgets (especially for high-dimensional problems), more efficient architectures and training objectives for relational inference, and tighter integration of quantized RL-tuned surrogates with edge devices to support real-time optimization in resource-constrained environments.

\bibliographystyle{IEEEtran}
\bibliography{bare_jrnl}

\end{document}